\documentclass{article}

\usepackage{arxiv}

\usepackage[utf8]{inputenc} 
\usepackage[T1]{fontenc}    
\usepackage{hyperref}       
\usepackage{url}            
\usepackage{booktabs}       
\usepackage{amsfonts}       
\usepackage{nicefrac}       
\usepackage{microtype}      
\usepackage{lipsum}

\usepackage{float}
\usepackage{color}
\usepackage{tabularray}
\usepackage{array}
\usepackage{multirow}
\usepackage{ragged2e}
\usepackage{booktabs}
\usepackage{vcell}
\usepackage{subcaption}
\usepackage{amsmath,amssymb,amsfonts}
\usepackage{algorithmic}
\usepackage{graphicx}
\usepackage{textcomp}
\usepackage{booktabs}
\usepackage{multirow}
\usepackage{subcaption}
\usepackage{caption}
\usepackage{siunitx}
\usepackage{pifont}
\usepackage{tipa}
\usepackage{tfrupee}
\usepackage{soul}
\usepackage{adjustbox}
\usepackage{tabularx,ragged2e,booktabs}

\usepackage{graphicx}
\graphicspath{ {./images/} }

\title{Countering Malicious Content Moderation Evasion in Online Social Networks: Simulation and Detection of Word Camouflage}

\author{ \href{https://orcid.org/
0000-0003-2165-0144}{\includegraphics[scale=0.06]{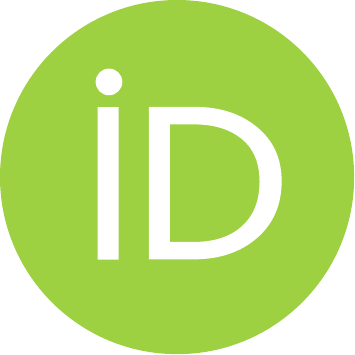}\hspace{1mm}{\'A}lvaro Huertas-Garc{\'i}a}\thanks{Use footnote for providing further
		information about author (webpage, alternative
		address)---\emph{not} for acknowledging funding agencies.} \\
	Department of Computer Systems Engineering\\
	Universidad Polit{\'e}cnica de Madrid\\
	Madrid, Spain \\
	\texttt{alvaro.huertas.garcia@upm.es} \\
	\And
	\href{https://orcid.org/
0000-0002-0800-7632}{\includegraphics[scale=0.06]{orcid.pdf}\hspace{1mm}Alejandro Mart{\'i}n} \\
	Department of Computer Systems Engineering\\
	Universidad Polit{\'e}cnica de Madrid\\
	Madrid, Spain \\
	\texttt{alejandro.martin@upm.es} \\
	\And
	\href{https://orcid.org/
0000-0003-4127-5505}{\includegraphics[scale=0.06]{orcid.pdf}\hspace{1mm}Javier Huertas-Tato} \\
	Department of Computer Systems Engineering\\
	Universidad Polit{\'e}cnica de Madrid\\
	Madrid, Spain \\
	\texttt{javier.huertas.tato@upm.es} \\	
	\And
	\href{https://orcid.org/
0000-0002-5051-3475}{\includegraphics[scale=0.06]{orcid.pdf}\hspace{1mm}David Camacho} \\
	Department of Computer Systems Engineering\\
	Universidad Polit{\'e}cnica de Madrid\\
	Madrid, Spain \\
	\texttt{david.camacho@upm.es} \\		
}

\begin{document}
\maketitle
\begin{abstract}
Content moderation is the process of screening and monitoring user-generated content online. It plays a crucial role in stopping content resulting from unacceptable behaviors such as hate speech, harassment, violence against specific groups, terrorism, racism, xenophobia, homophobia, or misogyny, to mention some few, in Online Social Platforms. These platforms make use of a plethora of tools to detect and manage malicious information; however, malicious actors also improve their skills, developing strategies to surpass these barriers and continuing to spread misleading information. Twisting and camouflaging keywords are among the most used techniques to evade platform content moderation systems. In response to this recent ongoing issue, this paper presents an innovative approach to address this linguistic trend in social networks through the simulation of different content evasion techniques and a multilingual Transformer model for content evasion detection. In this way, we share with the rest of the scientific community a multilingual public tool, named ``\textit{pyleetspeak}'' to generate/simulate in a customizable way the phenomenon of content evasion through automatic word camouflage and a multilingual Named-Entity Recognition (NER) Transformer-based model tuned for its recognition and detection. The multilingual NER model is evaluated in different textual scenarios, detecting different types and mixtures of camouflage techniques, achieving an overall weighted F1 score of 0.8795. This article contributes significantly to countering malicious information by developing multilingual tools to simulate and detect new methods of evasion of content on social networks, making the fight against information disorders more effective.

\end{abstract}


\section{Introduction}
\label{sec:introduction}

Regardless of the communication actions between users and their multimedia content, almost every social network today deals with malicious information (that is, misinformation, disinformation, misleading information or any other kind of information pollution) and the ostensible polarization of media discourse \cite{FaganFrank2020Osmc}. 


Content moderation has become one of the main ways social media platforms manage this situation. Moderation of content is critical to maintaining a safe and welcoming online environment. It involves reviewing and removing content that violates the rules and policies of a website or platform~\cite{content-filtering}. However, as these policies improve and new techniques are employed to monitor their compliance, so does how users can evade content moderation efforts~\cite{beyond-content-2018}. This effect entails severe consequences for both the platform and its users. If content moderation evasion is not addressed adequately, it can lead to the spread of harmful or illegal content, which can damage the reputation of the platform and put its users at risk~\cite{ads-in-osn, instagram-post-study}. Therefore, it is essential to provide these platforms with effective strategies to combat content moderation evasion.

In 2016, Facebook started an initiative against false claims pointing out those disproved by fact-checkers~\cite{mosseri_addressing_2016}. Later in 2020, with the emergence of the coronavirus, Twitter applied similar actions to manage the overabundance of dis/mis-information related to the COVID-19 pandemic, highlighting tweets that were considered to deliberately disseminate incorrect information to undermine public health~\cite{yoel__updating_nodate, SHAREVSKI2022102577, exsy-covid}. Additionally, since the origin of the pandemic, Twitter has facilitated developers and researchers access to their conversational content, providing a specific COVID-19 streaming endpoint~\cite{noauthor_covid-19_nodate} and an academic version of their API without timeline limitations~\cite{noauthor_twitter_nodate}, allowing a better study of the spread of hoaxes~\cite{martin2021factercheck}. Due to the same situation, the YouTube video platform recently established moderation policies for removing videos containing COVID-19 misinformation and limiting recommendations for anti-vaccination videos~\cite{noauthor_policy_YouTube}.

Undoubtedly, the COVID-19 pandemic has made evident the importance of fighting information disorders and moderating the content published on social networks. Moreover, content filtering has also received considerable critical attention in other fields such as terrorism, hate speech, misogyny, and sexism~\cite{beyond-content-2018, instagram-post-study}. For example, in 2017, Facebook, Google, Twitter, and Microsoft created the Global Internet Forum to Counter Terrorism (GIFCT) group, which collaborates with the European Commission to combat illegal online hate speech~\cite{AI-content-2020}. Instagram, Pinterest, and Tumblr are committed to limiting the results of searches with hashtags related to eating disorders and sexual abuse, among others~\cite{beyond-content-2018, instagram-post-study}.


Nevertheless, malicious actors present on these platforms are aware of these content-moderation rules. Recently, to evade this content filtering, it has been demonstrated that these actors twist and camouflage key parts of speech using different techniques such as leetspeak~\cite{instagram-post-study, bridge_gap_leet_2005} (which involves generating visually similar character strings by replacing alphabet characters with other symbols), word inversion or inserting punctuation characters into words,  procedures belonging to the named word camouflaging~\cite{romero_wordcamo_2021}. These evasion methods threaten the capabilities of the platform's systems and allow malicious actors to continue spreading falsehood, misleading, and harmful content, as shown below in Section~\ref{sec:evasion-examples}.

The primary purpose of this study is to provide new instruments in response to these rapid changes in content moderation evasion. The following are the new contributions of our research concerning previous studies.

\begin{itemize}

    \item We present and share with the rest of the scientific community a novel methodology to generate/simulate the content evasion phenomenon from a multilingual level in a customizable way. For the sake of reproducibility, the methodology approach is presented as a public Python package named ``pyleetspeak''\footnote{\label{foot:pyleetspeak}\href{https://pypi.org/project/pyleetspeak/}{\url{https://pypi.org/project/pyleetspeak/}}}. It should be noted that the tool is customizable and is not language dependent.

    \item We present a curated synthetic multilingual dataset\footnote{\label{foot:link-data}\href{https://github.com/Huertas97/XX_NER_WordCamouflage}{\url{https://github.com/Huertas97/XX_NER_WordCamouflage}}} of camouflaged words with applied quality filters. The dataset is shared in different formats to facilitate its applicability to other researchers. The languages considered are English, Spanish, French, Italian, and German.

    \item We derive a multilingual Transformer-based model\footnote{\label{foot:model}\href{https://huggingface.co/Huertas97/xx_LeetSpeakNER_mstsb_mpnet}{\url{https://huggingface.co/Huertas97/xx_LeetSpeakNER_mstsb_mpnet}}} to detect and discern different word camouflage techniques to prevent content evasion.    

    \item We evaluated the potential of multiclass camouflage Named Entity Recognition at the multilingual level by comparing the developed multilingual model with monolingual baseline models for the different languages considered.

    \item Finally, we continue previous research~\cite{aida_model_2021} by reaffirming the usefulness of multilingual pre-training in semantic similarity (mSTSb) to increase the generalizability of multilingual models.

\end{itemize}

This paper is organized as follows. Section~\ref{sec:literature} begins by examining previous work on content moderation and content moderation evasion techniques. Section~\ref{sec:material} introduces our new methodology to simulate content moderation evasion and generate annotated data. Section~\ref{sec:experiment-setup} explains the data used to generate word camouflage and train multilingual word camouflage NER models. It also explains the experimental setup for the development of Transformer-based models for detecting word camouflaging. The experimental results are discussed in Section~\ref{sec:results}. Finally, our conclusions are drawn in the final section~\ref{sec:conclusions}.

\section{Literature Review}
\label{sec:literature}

\subsection{Content moderation}

Content moderation is the process of screening and monitoring user-generated content online to suppress communications that are deemed undesirable. As Gerrard \& Thornham~\cite{content_sex_moderation_2020} have highlighted at present, there are two dominant forms of social media content moderation: automated and human. However, the growing amount of content uploaded to social media platforms makes it impossible to rely exclusively on the human content moderation approach. 

Traditional practices to limit the disruption that can be caused by antisocial behavior consist of blocking messages based on basic text properties (e.g., length), interaction parameters (e.g., posting frequency, reply frequency), or according to the standards of designated moderators~\cite{trad_moderation_004}. These practices had some drawbacks, such as their applicability to small, medium-sized conversations. These initial practices led to more sophisticated and scalable forms of automated moderation~\cite{Elkin-KorenNiva2020CaRt, cobbe_algorithmic_2021}.

Although these systems from online content platforms remain opaque and poorly understood, two different methods are known. In some cases, automated systems employ fingerprinting or hash matching techniques to compare new content against known data and databases of unwanted or flagged content~\cite{ content_sex_moderation_2020, sumpter_outnumbered_2018}. In other cases, automated systems will be machine learning systems trained on large datasets to spot new or previously unseen allegedly illegal material and remove it, block it, or filter it, which can also be used to create more training data or assist in human moderation~\cite{Elkin-KorenNiva2020CaRt, cobbe_algorithmic_2021, ofcom_use_2019}. Consequently, these automated content filtering methods benefit from data sharing. For example, the four members of GIFCT share best practices and databases to develop their automated systems. An example of the importance of this data sharing is the Christchurch attack in New Zealand in 2019, where a terrorist broadcast the murder of more than 50 people on Facebook live stream. After that incident, Facebook shared this information with other platforms. Every video or image uploaded by ordinary users from any of these platforms would now be checked against it to check whether it should be blocked or not~\cite{content_sex_moderation_2020, sumpter_outnumbered_2018}.


In the literature, many examples illustrate the significant role that automated systems are already playing in content moderation. According to~\cite{youtube_ban_results}, 98\% of the videos YouTube removes for violent extremism are flagged by machine learning algorithms and help human reviewers remove videos nearly five times faster. Moreover, the visualization time for harmful antivaccine content videos has been reduced by more than 70\% after YouTube limited the recommendation for such videos~\cite{ferreira_antivaccine_2020}. In 2021, Twitter temporarily locked Donald Trump's account for allegedly inciting an attack on the United States Capitol and permanently suspended it for violating Twitter's Glorification of Violence guidelines~\cite{_twitter_inc_permanent_2021}. Twitter has also been testing features to allow people to report potentially misleading information, and has recently expanded to Brazil, the Philippines, and Spain~\cite{twitter_blog_nuevo_2022}. Additionally, in 2021, Facebook has started a campaign to explicitly ban ``any content that denies or distorts the Holocaust''~\cite{_bickert_removing_2020}. Similarly, other researchers have also developed machine learning solutions to help security administrators detect phishing content in emails and social networks~\cite{phising_detection_ml}. Other researchs~\cite{wall_filter, user_wall_2} have explored the application of machine learning classifier on  online social networks to facilitate content-based filtering assistance to avoid unwanted content displayed on the user wall. 

These situations and incidents like those raised during the COVID-19 pandemic clearly show that automated moderation systems have become necessary to manage growing public expectations for greater responsibility, safety, and security on the platforms~\cite{AI-content-2020}. Nevertheless, content filtering automated systems depend on their ability to analyze the material uploaded, potentially vulnerable to recent content evasion techniques, such as word camouflaging.


\begin{table}
\centering
\caption{Examples of Leetspeak technique applied in different situations}
\label{table:leet-example}
\begin{tblr}{
  width = \linewidth,
  colspec = {Q[m,c,10em]Q[m,10em]Q[m,12em]Q[m,c,10em]},
  hline{2-6} = {-}{},
}
\textbf{Case of study }               & \textbf{ Original }          & \textbf{Camouflaged}                                                              & \textbf{Source}                                                  \\
Gaming                                & {noobs\\owned\\skiils\\fear} & {n00bz\\pwn3d\\sk11lz\\ph34r}                                                     & Blashki et al. 2005                                              \\
Password                              & HBOpassword                  & \#B0p4\$\$w0r\textbar{})                                                          & Hong et~al. 2021                                                 \\
Cybersquatting                        & incibe.es                    & {incive.es\\incIbe.es\\inci-be.es\\inicbe.es}                                     & {Instituto Nacional\\de \\Ciberseguridad \\~ ~ ~ (INCIBE)~ ~ ~~} \\
{Social Media \\COVID-19 \\Infodemic} & {vacuna\\\\\\\\\\covid}      & {v4cun4\\b4cun4\\v@(u¬a\\nacuva\\V.A.C.U.N.A\\\\k0 b1t\\K0b1d\\c0*vid\\C(o(v(i(d} & ~ ~ EU Disinfolab~ ~ ~                                           
\end{tblr}
\end{table}

\subsection{Content Moderation Evasion Techniques}
\label{sec:evasion-examples}

Before proceeding further, we would like to provide a sensitive content disclaimer to warn the reader that there is a possibility of finding offensive content in the examples included in the work presented.

Community-driven Internet spaces, especially social networks, have always presented unique dialects and slang terminology~\cite{Blashki2005GameGG, Shaari2015NetspeakAA}. These ever-changing dialects are the natural result of short codes to facilitate communication, user interactions on social networking platforms, and the adaptation of language to new technologies~\cite{bridge_gap_leet_2005}. However, the emergence of new terms or dialects can result from intentionally camouflaging messages without impacting the information transmitted to avoid content moderation.

One of the main techniques for this purpose is \textit{leetspeak}. \textit{Leetspeak} is a written language in which characters are changed to other characters or combinations of characters that visually resemble the original~\cite{Blashki2005GameGG, passwords_leet_2019} (see Table \ref{table:leet-example}). Even though there is still a great deal of uncertainty about its origin, there is no doubt that its initial use was related to content evasion.

One reliable hypothesis~\cite{Blashki2005GameGG} holds that hackers initially used leetspeak in the early stages of the Internet to prevent their content from being accessible. At that time, most search systems searched for keywords in the text to recommend relevant content, and users who were reluctant to share their information substituted certain letters in words to avoid being included in searches. Other observations~\cite{fuchs__2013} indicate that leetspeak' origins can be found on Bulletin Board Systems, much like today's forums,  to avoid censorship measures present in instant messaging systems. 

There is also controversy as to why the term leetspeak was coined. In their analysis of the adaptation of language by a community of young people who play computer games, Blashki et al. ~\cite{Blashki2005GameGG} proposed that the root term ``leet'' was originated from 31337 ``eleet'', the UDP port used by a hacker group to access Windows 95 using the Back Orifice hacking program. Other researches~\cite{fuchs__2013} propose that it was first considered ``elite'' as only a few people could encode and decode it, therefore using the term ``leet'' to refer to this particular group. Subsequently, the use of leetspeak became more popular, and it became integrated into the gaming community, particularly by Counter Strike and World of Warcraft players~\cite{Blashki2005GameGG, fuchs__2013}.


\makeatletter
\DeclareRobustCommand{\PHP}{%
  \begingroup
  \leavevmode\,\vphantom{P}%
  \dimen\z@=.5\fontcharht\font`P\relax
  \dimen\tw@=0.33333\dimen\z@
  \ooalign{%
    \raisebox{\dimexpr\dimen\z@+2\dimen\tw@-0.4pt}{\rule{\fontcharwd\font`P}{0.4pt}}\cr
    \raisebox{\dimexpr\dimen\z@+\dimen\tw@-0.2pt}{\rule{\fontcharwd\font`P}{0.4pt}}\cr
    P\cr
  }%
  \,\endgroup
}
\makeatother

\newcolumntype{L}{>{\RaggedRight\arraybackslash}X}
\newcolumntype{Y}{>{\centering\arraybackslash}X}

\begin{table*}[tpb]
\caption{Examples of word camouflage using different methods from pyleetspeak.}
\label{table:pyleet_examp}
\centering
\bgroup
\def\arraystretch{1.25}%
\begin{tabularx}{\textwidth}{@{}Y|YYY|c|Y@{}}
\toprule
\multicolumn{1}{c}{\multirow{2}{*}{\textbf{Word }}} & \multicolumn{3}{c}{\textbf{Leetspeaker}}                                                                                                                      & \multicolumn{1}{c}{\multirow{2}{*}{\textbf{Punctuation }}} & \multicolumn{1}{Y}{\multirow{2}{*}{\textbf{Inversion }}}  \\ 
\multicolumn{1}{c}{}                                & \multicolumn{1}{c}{\textbf{Basic}}                                        & \multicolumn{1}{c}{\textbf{Intermediate}} & \multicolumn{1}{c}{\textbf{Advanced}} & \multicolumn{1}{c}{}                                       & \multicolumn{1}{c}{}                                      \\ 
\hline
Vacuna                                               & \begin{tabular}[c]{@{}c@{}}V@c\_n@\\V$\Delta$c\"{u}n$\Delta$\end{tabular}             &     \begin{tabular}[c]{@{}c@{}}\textbackslash{}/4[u\textbar{}\textbackslash{}\textbar{}4\\V.q\"{u}n$\Delta$\end{tabular}                   
&    \begin{tabular}[c]{@{}c@{}}\textbar{}/a[L\textbar{}/\textbackslash{}/a\\\textbackslash{}/4[V\textbar{}\textbackslash{}\textbar{}/\textbackslash{}\end{tabular}                                 &       \begin{tabular}[c]{@{}c@{}}      'V'a'c'u'n’a\\Vac'u=na\end{tabular}                                               &       nacuVa
                                                \\
\hline
Covid                                                & \begin{tabular}[c]{@{}c@{}}C0v1d\\C\o{}v\textexclamdown{}d\end{tabular}              &  \begin{tabular}[c]{@{}c@{}} k.vb!t\\ C0\ding{116}!t\end{tabular} 
                                & \begin{tabular}[c]{@{}c@{}} {[}ov\textbar{}d \\ C[]\textbackslash{}\textbar{}!\textbar{}\textgreater{} \end{tabular}                                      &   \begin{tabular}[c]{@{}c@{}} ?C?o?v?i?d \\ 'C-ovid \end{tabular}                                                      &    vidCo                                                       \\
\hline
Plandemia                                            & \begin{tabular}[c]{@{}c@{}}Plnd\textsterling{}m* \\ Pland.mi$\Delta$\end{tabular}         &  \begin{tabular}[c]{@{}c@{}}  \PHP la¬t\mbox{\texteuro}{}mla \\ P$1\Delta\pi$d3\textturnw{}ia \end{tabular}
                                       &  \begin{tabular}[c]{@{}c@{}} \textbar{}\textgreater{}landem[]/\textbackslash{} \\ Pl/\textbackslash{}nde[V]1a \end{tabular}                                    &   \begin{tabular}[c]{@{}c@{}}    !P!l!a!n!d!e!m!i!a  \\ Plan/demi/a \end{tabular}                                                     &         dePlanmia                                                  \\
\hline
Inmigrant                                            & \begin{tabular}[c]{@{}c@{}}1nmigr\_nt \\ Inm*grnt\end{tabular}        &   \begin{tabular}[c]{@{}c@{}}  I$\pi$mig\rupee$\Delta$nt \\ In\textturnw{}\textexclamdown{}g\rupee{}a$\pi$t \end{tabular}
                    &  \begin{tabular}[c]{@{}c@{}} In[]V[]igr/\textbackslash{}nt \\ In\^{}\^{}][(\_+ra\textbar{}\textbackslash{}\textbar{}t \end{tabular}                                 & \begin{tabular}[c]{@{}c@{}}   .I.n.m.i.g.r.a.n.t\\+I+nmi+gran\end{tabular}                                                        &           migrantIn                                                \\
\hline
Dictatorship                                         & \begin{tabular}[c]{@{}c@{}}Dict*t*rship \\ Dict$\Delta$t0rship\end{tabular} &        \begin{tabular}[c]{@{}c@{}}Dict.\st{T}\o{}rs\#i\PHP \\  Di\textcopyright{}tat*rz\#lp\end{tabular}                        &    \begin{tabular}[c]{@{}c@{}}Ditat(0)/2ship\\(\textbar{}ict/\textbackslash{}t$<>$rship\textbar{}7\end{tabular}                                &        \begin{tabular}[c]{@{}c@{}}Dicta:torsh:ip \\ D\textbar{}i\textbar{}c\textbar{}t\textbar{}a\textbar{}t\textbar{}o\textbar{}r\textbar{}s\textbar{}h\textbar{}i\textbar{}p\end{tabular}                                                &                                  Dicortatship                         \\
\hline
Genocide                                             & \begin{tabular}[c]{@{}c@{}}Gen\o{}cld \\ G\%nocide\end{tabular}         &            \begin{tabular}[c]{@{}c@{}}        G3$\pi$o\textcopyright{}ite \\ Gen\textcopyright{}id@ \end{tabular}                   &   \begin{tabular}[c]{@{}c@{}}  9e\textbar{}\textbackslash{}\textbar{}o[i[)e \\ Gen$<>$ci\textbar{}$>$e \end{tabular}                                    &    \begin{tabular}[c]{@{}c@{}}G;enoc;i;de \\ G=e=n=o=c=i=d=e\end{tabular}          
&     oGencide
                                                      \\
\bottomrule
\end{tabularx}
\egroup
\end{table*}

Nowadays, leetspeak is mainly associated with online multiplayer gamers~\cite{Shaari2015NetspeakAA}. Interestingly, the leetspeak camouflaging technique is also explored in the field of password generation and password security. Golla et al.~\cite{password_leet_use_2016} mention the practice of using leetspeak to modify characters in passwords to make them more secure, but at the same time facilitate remembering them~\cite{passwords_leet_2019}. Passwords replaced with leetspeak have been tested with password strength estimators, such as \textit{zxcvbn}~\cite{password_dropbox_2016}, and get a high-security rating, as they combine alphabet characters, numbers, and special symbols~\cite{password_leet_2021}. In the same way, leetspeak has been related to cybersquatting, the registration of a domain name that is the trademark of another~\cite{noauthor_cybersquatting_2019} (see Table \ref{table:leet-example}).

Remarkably, the work of Peng et al.~\cite{spam_leet_2018} reveals the vulnerability of machine learning algorithms in detecting spam in emails when they included leetspeak. Because leetspeak uses unconventional spelling and punctuation, the revealed the difficulty for the automatic systems to accurately identify and interpret the words and phrases being used. Hence, this can make it easier for attackers to evade detection and spread harmful or illegal content. To address this vulnerability, in this works we develop Transformer-based models trained on a wide range of text inputs, including examples of word camouflage, to improve their accuracy and ability to identify this type of language.

Previous studies~\cite{slang_media_2016, slang_media_2020} have analyzed a variety of slangs, including leetspeak, in social media, but since the emergence of coronavirus in 2019, this situation has become more pronounced, with leetspeak being a way to circumvent censorship. In~\cite{romero_wordcamo_2021} the authors analyzed how malicious actors camouflage virus-related Spanish keywords to spread misleading content, revealing their skills in developing new techniques to continue spreading their message and the complexity of tackling this phenomenon (see Table \ref{table:leet-example}).

Finally, evidence of the presence of these methods of content evasion in social networks can be clearly observed through a comparison of the search results for terms associated with hateful behaviour and which do not comply with the Community Guidelines in original and camouflaged formats. 


The term ``self-harm'' on the TikTok platform is associated with the health and well-being of people and is part of their community guidelines for safety. Searching for this term redirects to an official contact for hope support\footnote{\href{https://www.tiktok.com/search?q=self-harm}{\url{https://www.tiktok.com/search?q=self-harm}}}. However, if we change the search term to ``s3lf-harm'' we bypass moderation and access videos with potential content depicting, promoting, normalizing or glorifying activities that could lead to suicide or self-harm\footnote{\label{foot:sel-harm}\href{https://www.tiktok.com/search?q=s3lf-harm}{\url{https://www.tiktok.com/search?q=s3lf-harm}}}. Other examples are the term ``incel'', which in an extreme way refers to those who have violent or hateful attitudes towards women or towards society in general~\cite{Moskalenko_González_Kates_Morton_2022}. This term does not return any results and redirects us to the community guidelines\footnote{\href{https://www.tiktok.com/search?q=incel}{\url{https://www.tiktok.com/search?q=incel}}}. Again, we can bypass this moderation and access content by using ``1ncel'' instead\footnote{\label{foot:incel}\href{https://www.tiktok.com/search?q=1ncel}{\url{https://www.tiktok.com/search?q=1ncel}}}. 
The same happens in other languages. For example, the search for the Spanish terms ``violación'' and ``pornografía'', in English ``rape'' and ``pornography'', does not return any results\footnote{\href{https://www.tiktok.com/tag/violacion}{\url{https://www.tiktok.com/tag/violacion}}}\footnote{\href{https://www.tiktok.com/search?q=pornografía}{\url{https://www.tiktok.com/search?q=pornografía}}}, unlike their camouflaged version ``v1olaci0n''\footnote{\label{foot:violacion}\href{https://www.tiktok.com/tag/v1olaci0n}{\url{https://www.tiktok.com/tag/v1olaci0n}}} and ``p0rnograf!a''\footnote{\label{foot:porno}\href{https://www.tiktok.com/search?q=p0rnograf!a}{\url{https://www.tiktok.com/search?q=p0rnograf!a}}} with content with more than 8 million views. 

As a result of these findings, developing tools that mimic and detect content avoidance techniques are essential for content moderation in the fight against information disorders.



\section{Methodology}
\label{sec:material}


This section shows the methodology for generating camouflage data that will be used to train and evaluate models in the detection of camouflaged words in content evasion. Firstly, we describe how camouflage techniques are simulated by developing a public Python package that applies the rules described in the literature. 
Secondly, we explain how these camouflaging techniques are applied to an input text to obtain a camouflaged version of the input text with NER annotation to obtain training and evaluation data for the models. 

\subsection{Simulation of Word Camouflage Techniques}
\label{sec:source-data}

Taking into account the importance of linguistic characters used in a language to generate camouflaged versions, it is necessary to point out that the tools developed in the package are multilingual (+20 languages\footnote{ar, az, da, de, el, en, es, fi, fr, hu, id, it, kk, nb, ne, nl, pt, ro, ru, sl, sv, tg, tr}). This tool has been tested in English, Spanish, French, Italian, and German. However, it can be easily extensible to the rest of Latin-derived alphabet languages (i.e., most of the Western European languages). As a reminder, the tool described below is publicly available in the Python ``pyleetspeak'' package\textsuperscript{\ref{foot:pyleetspeak}}. 


We have designed three approaches to emulate content evasion strategies based on text camouflage modification. These methods were developed in relation to the results described in the analysis by Romero-Vicente et al.~\cite{romero_wordcamo_2021} of recent content avoidance techniques on social networks: LeetSpeaker, PuntctuationCamouflage and InversionCamouflage modules.

    
    

\subsubsection{LeetSpeaker module}

This module applies the well-known \textit{leetspeak} method to produce visually similar character strings by replacing alphabet characters with special symbols or numbers. There are many ways to use leetspeak, from basic vowel substitutions to advanced combinations of various punctuation marks and symbols.

The \textit{leetspeak} alphabet in its simplest form substitutes vowels, but it can be pretty complex when substituting consonants as well. As a consequence, the leetspeak modifications implemented in the package are organized into five different modes depending on the visual complexity of the camouflage. The implemented changes, available in our repository\footnote{\href{https://github.com/Huertas97/pyleetspeak/blob/main/pyleetspeak/modes.py}{\url{https://github.com/Huertas97/pyleetspeak/blob/main/pyleetspeak/modes.py}}}, have been obtained from different sources~\cite{bridge_gap_leet_2005,
romero_wordcamo_2021,
Blashki2005GameGG, 
passwords_leet_2019, 
fuchs__2013, 
password_dropbox_2016, craenen_leet_cheatsheet}. Nevertheless, the tool is flexible, as new possible substitutions can be specified for its adaptability to new unexplored or ever-evolving scenarios. 

Similarly, in the case of LeetSpeaker, other parameters that can be set to customize the camouflage result are the probability of changing a character type (e.g., change ``a" for ``@") and the frequency of substitution, named \textit{chg\_prb} and \textit{chg\_frq} in the package, respectively. The frequency of substitution refers to the number of positions to change among all matches of the same character (e.g., whether to replace the two ``a" letters with the ``@" symbol in ``vaccination"). In the case of an original character that has more than one possible substitution, one is randomly selected from a uniform probability distribution. 

In a real scenario, such as social networks, users tend to use the same type of substitution for all occurrences of the same character. To emulate this situation, it is possible to define whether or not the transformations performed by LeetSpeaker and PunctuationCamouflage should be independent of each other using the \textit{uniform\_change} parameter. In other words, it determines whether the same substitution character should be used in all positions where the original text is modified. For example, if ``a" can be replaced by ``@" or ``4", select whether ``vaccination" should be ``v4ccin4tion" or ``v@ccin4tion".

\subsubsection{PuntctuationCamouflage module}

Another method to create visually similar character strings is to insert punctuation symbols into the text (see Table \ref{table:pyleet_examp}).

Regarding PunctuationCamouflage module, it can be further customized to inject punctuation symbols in hyphenate locations (i.e., syllables) or between any character. In addition, the number of punctuation symbols to inject can be specified. Interestingly, Romero-Vicente et al.~\cite{romero_wordcamo_2021} reported that malicious actors usually use punctuation camouflaging, inserting punctuation symbols between all letters of keywords (e.g., ``C.O.V.I.D.-1.9''). This behavior can also be reproduced in pyleetspeak without previously specifying the input text length to be camouflaged using the \textit{word\_splitting} parameter. The default punctuation symbols applied come from the built-in Python ``string" module, but the user can specify the symbols to use.

\subsubsection{InversionCamouflage module}

Although not as common as the previous methods, word inversion can also be used to confuse moderating algorithms. For this reason, InversionCamouflage module creates new camouflaged versions of words by inverting the order of the syllables (see Table \ref{table:pyleet_examp}).

Word inversion is implemented by detecting the syllables that constitute a word. Once the word has been separated into its syllables, two of these syllables are randomly selected and inverted with respect to each other. As in the previous methods, the inversion is customizable and it is possible to indicate the maximum distance between two syllables to be interchanged. If there are several possibilities for syllable interchange, one is chosen at random.

Further details on the adopted methodology can be found in our repository. Similarly, this tool can be tested in the demo application Leetspeaker App\footnote{\href{https://github.com/Huertas97/LeetSpeaker_App}{\url{https://github.com/Huertas97/LeetSpeaker_App}}} developed with Dash \cite{plotly}, and other examples of the package ``pyleetspeak'' package are shown in Table \ref{table:pyleet_examp}.

\begin{figure*}[t]
\centering
\includegraphics[width=\linewidth]{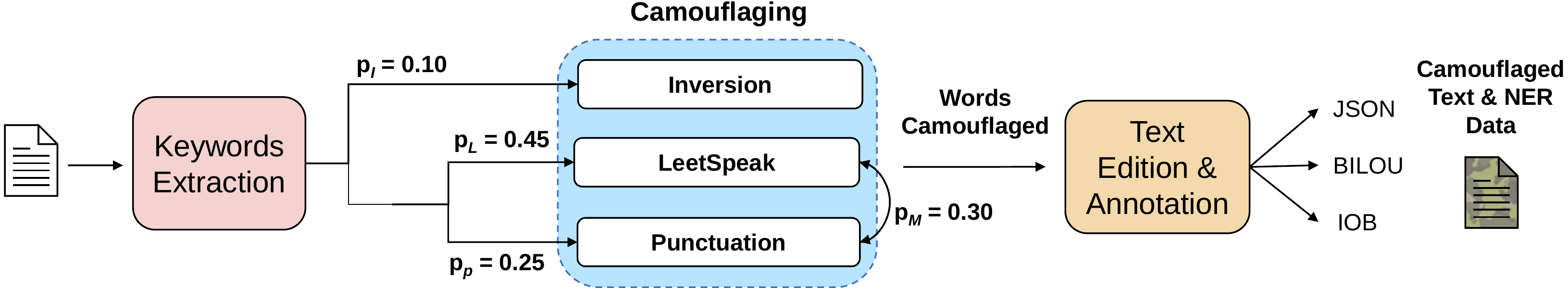}
\caption{Named Entity Recognition data generation diagram. $p_L$, $p_I$ and $p_P$ represent the probability of applying leetspeak, inversion, and punctuation camouflage techniques. $p_M$ represents the probability of applying a mixture of different techniques once a technique has been previously applied.}
\label{fig:diagram_methodology}
\end{figure*}

\subsection{Word camouflage NER data generator}\label{sec:ner-generator}

As depicted in Figure \ref{fig:diagram_methodology}, ``pyleetspeak'' package transforms an input text into a camouflaged version. The use of word camouflaging usually involves changing the most critical words of a sentence instead of leetspeak all the words in the text. Thus, KeyBERT~\cite{grootendorst2020keybert} is used to extract the most semantically relevant words and apply them different word camouflaging methods presented above. Finally, the camouflaged entities in the output text are annotated in Spacy format~\cite{montani_spacy_2020}. 

KeyBERT incorporates state-of-the-art Transformer models for keyword extraction~\cite{martin2021factercheck}. This method represents a viable alternative to traditional statistical methods for keyword extraction, as it benefits from the use of powerful Transformer-based models~\cite{vaswani2017attention, devlin2019bert}. These state-of-the-art models have recently radically transformed the Natural Language Processing area for their ability to generate powerful semantically aware text representations. Precisely, KeyBERT exploits this semantic awareness to compute words and text embeddings, then extracts the most semantically relevant words of a text using cosine similarity as similarity function. Consequently, the most similar words are the keywords that best describe the meaning of the text. By incorporating KeyBERT into our NER data generator, we better meet the expectations of real-world scenarios in which malicious actors camouflage vital concepts in conversation to evade content moderation. Additionally, we employ the \textit{mstsb-paraphrase-multilingual-mpnet-base-v2} fine-tuned on the multilingual Semantic Textual Similarity Benchmark~\cite{aida_model_2021} as the Transformer model for keyword extraction. However, any other HuggingFace~\cite{wolf-etal-2020-transformers} model can be selected. The tool has been optionally adapted to always incorporate user-specified keywords to better control the camouflaged output.

Finally, the camouflaged NER annotated data generated is composed of 4 different types of entities. LEETSPEAK, PUNCT\_CAMO, INV\_CAMO represent the different camouflage methods implemented, and the MIX entity, which represents the combination of leetspeak and punctuation camouflage. It is also worth noting that, in order to increase the interpretability of the process, besides the annotated camouflage data, the tool returns a dictionary containing the parameters applied to each instance (e.g., keywords extracted, type of camouflage applied, values of the parameters).

\renewcommand{\arraystretch}{1.2}

\begin{table}
  \fontsize{8}{8}\selectfont 
\centering
\caption{Breakdown of the multilingual data corpus after quality filtering according to each language, resource and division to develop NER word camouflage models.}
\label{table:data-breakdown}
\begin{tblr}{
  width = \textwidth,
  colspec = {Q[195]Q[60]Q[52]Q[50] Q[60]Q[52]Q[50] Q[60]Q[52]Q[50] Q[60]Q[52]Q[50] Q[60]Q[52]Q[50]},
  rowsep=5pt,
  cells = {c}, 
  column{1} = {l},
  vline{5,8,11,14} = {-}{},
  hline{3,7} = {-}{},
  hline{8} = {-}{},
}
                   &       & \textbf{EN} &      &       & \textbf{ES} &      &       & \textbf{FR} &      &       & \textbf{IT} &      &       & \textbf{DE} &      \\
                   & Train & Dev         & Test & Train & Dev         & Test & Train & Dev         & Test & Train & Dev         & Test & Train & Dev         & Test \\
{\begin{tabular}[c]{@{}c@{}}News Commentary\end{tabular}} & 645   & 73          & 80   & 11341 & 1255        & 1398 & 589   & 69          & 72   & 130   & 15          & 16   & 859   & 94          & 103  \\
ParaCrawl          & 19678 & 2180        & 2445 & 23113 & 2580        & 2853 & 21424 & 2373        & 2656 & 22279 & 2474        & 2772 & 20813 & 2334        & 2563 \\
TED2020            & 15548 & 1712        & 1907 & 15717 & 1758        & 1938 & 14225 & 1580        & 1751 & 15091 & 1680        & 1879 & 14742 & 1667        & 1818 \\
WikiMatrix         & 15557 & 1707        & 1927 & 15529 & 1717        & 1903 & 14489 & 1626        & 1806 & 15124 & 1662        & 1889 & 14457 & 1618        & 1782 \\
\textbf{Total}     & 51428 & 5672        & 6359 & 65700 & 7310        & 8092 & 50727 & 5648        & 6285 & 52624 & 5831        & 6556 & 50871 & 5713        & 6266 
\end{tblr}
\end{table}

\section{Experimental Setup}
\label{sec:experiment-setup}

This section presents the collected multilingual non-camouflaged text data, how they are camouflaged to simulate content evasion, and finally, the monolingual and multilingual models trained for NER word camouflage detection. 




\subsection{Non-Camouflaged Training Data}
\label{training-data}

To the best of our knowledge, no dataset with annotated word camouflage modifications is available to train and evaluate our models. Although there are some existing datasets in the field, none of them meet our specific needs and, therefore, we cannot use them directly. In fact, one of the main contributions of our work is precisely the creation of a new publicly available dataset\textsuperscript{\ref{foot:link-data}} that fills the existing gap and that we hope will be useful to other researchers in the future.

The dataset with synthetic camouflaged words is elaborated from non-camouflaged texts since, to train the models for word camouflage detection, the camouflage modifications present in the text must be previously annotated. Therefore, we employ non-camouflaged datasets, as it allows us to control that the camouflaging belongs exclusively to the modifications derived by our word camouflage generator tool.

We have chosen the following resources due to the variety of text types among these resources:

\begin{itemize}

    \item \textbf{OPUS News-Commentary}~\cite{opus_news_data}: A parallel corpus of political and economic news commentaries in 12 languages was crawled from the web site Project Syndicate provided by WMT.

    \item \textbf{OPUS ParaCrawl}~\cite{tiedemann-2012-parallel}: Multilingual parallel corpora from around 150k website domains and across 23 EU languages collected in the ParaCrawl project~\cite{banon-etal-2020-paracrawl} cofinanced by the European Union.

    \item \textbf{TED2020}~\cite{reimers_2020_multilingual_sentence_bert}: This dataset contains a crawl of nearly  \num[group-separator={,}]{4000} TED and TED-X transcripts from July 2020. The transcripts have been translated by a global community of volunteers into more than 100 languages.

    \item \textbf{WikiMatrix}~\cite{schwenk2019wikimatrix}: Mined parallel sentences from the content of Wikipedia articles in 85 languages. In this project, a 1.04 score threshold was used for parallel text extraction.     
\end{itemize}


The languages considered in this work are English, Spanish, French, Italian, and German. For each language, data is extracted from the different resources shown above, discarding those texts with a length of less than 3 characters. Subsequently, the extracted data are camouflaged, discarding the annotated data that do not pass the Spacy quality filter\footnote{\label{foot:spacy-filter}\href{https://spacy.io/api/cli\#debug-data}{\url{https://spacy.io/api/cli\#debug-data}}} and are split in a stratified way into train (81\%), validation (9\%) and test (10\%) sets to perform training and evaluation of multilingual and monolingual NER models.  The breakdown of the final data considered in this work according to the language and type of resource can be found in Table~\ref{table:data-breakdown} and is publicly available on GitHub\textsuperscript{\ref{foot:link-data}}.

The parameters used to camouflage the data with the camouflage tool previously presented and how the quality of the data is evaluated with the Spacy tool will be explained in Subsection~\ref{sec:ner-data} below.


\subsection{Annotated NER data: Camouflaging parameters and Quality filter}\label{sec:ner-data}

To carry out camouflage and annotation of the modified words in the data shown in the previous section, we assigned different occurrence probabilities to the camouflage methods using the methodology presented in Section \ref{sec:material}. 
As shown in Figure~\ref{fig:diagram_methodology}, in 10\% of the cases, word inversion is employed. In the remaining cases, leetspeak is applied with 45\% probability, 25\% punctuation camouflage, and 30\% combination of both. Although the use of inversion modification in conjunction with other camouflage techniques is potentially available, we have not found any evidence to support this approach. Therefore, in our work, we have applied inversion modification in a stand-alone mode and  the camouflaging techniques in a distribution that we believe best reflect the reality, although they are fully customizable.  

For the sake of reproducibility, all the values used in the various parameters to generate the NER data are shown in Table \ref{table:parameter_NER}. To ensure that the different train, validation and test splits have the same distribution of entity types, the parameters of the NER data generator are equally employed across languages to create the camouflaged data version.  

Upon obtaining the modified datasets, the Spacy data debugger\textsuperscript{\ref{foot:spacy-filter}} is applied. This quality filter removes possible duplicates from the source data, checks that there are no overlaps between the training and evaluation data,  that there are a good number of examples for all labels, there are examples with no occurrences available for all labels, there are no entities consisting of or starting/ending with blanks, and there are no entities crossing sentence boundaries. 

Finally, the camouflaged annotated data obtained are saved in Spacy formats. The customizable ``pyleetspeak'' tool presented also provides a format converter to transform Spacy NER format to JSON, BILUO, or IOB format, the one used by the Hugging Face community. 

The curated synthetic multilingual dataset obtained is available on GitHub\textsuperscript{\ref{foot:link-data}} in the different formats indicated above.

\begin{table}[t!]
\centering
\begin{minipage}[h]{0.52\textwidth}

\centering
\captionof{table}{Summary of parameter values used to camouflage and annotate text data to develop NER models.}
\label{table:parameter_NER}
\resizebox{\columnwidth}{!}{%
\begin{tabular}{ll} \toprule
LeetSpeaker                                                     & \begin{tabular}[c]{@{}l@{}}change\_prb = 0.8\\change\_frq = 0.5\\probability basic modes = 0.5\\probability intermediate modes = 0.4 \\probability advanced mode = 0.1\\probability uniform\_change = 0.6\end{tabular}  \\ \hline
\begin{tabular}[c]{@{}l@{}}Punctuation\\Camouflage\end{tabular} & \begin{tabular}[c]{@{}l@{}}probability hyphenation = 0.5\\probability uniform\_change = 0.6\\probability word\_spliting = 0.5\\number injections = randint(1, lenght)\end{tabular}                                      \\ \hline
\begin{tabular}[c]{@{}l@{}}Inversion\\Camouflage\end{tabular}   & max distance = randint(1, 4)                                                                                                                                                                                            \\ \hline
\multirow{3}{*}{KeyBERT}                                        & \begin{tabular}[c]{@{}l@{}}model =\\mstsb-paraphrase-multilingual-mpnet-base-v2\end{tabular}                                                                                                                            \\
                                                                & max number of keywords = 5~~                                                                                                                                                                                            \\
                                                                & keywords n\_gram = (1, 1)                                                                                                                                                                                               \\ \bottomrule
\end{tabular}
}

\end{minipage}
\hspace{1mm}
\centering
\begin{minipage}[h]{0.44\textwidth}
  \centering
\captionof{table}{Parameters considered during the training of the models for word camouflaged Named Entity Recognition wit Spacy}\label{table:training-params}
\resizebox{\columnwidth}{!}{%
\begin{tabular}{cl} \toprule
\multirow{4}{*}{learning rate}                          & \textcolor[rgb]{0.102,0.11,0.122}{initial\_rate =~}\textcolor[rgb]{0.102,0.11,0.122}{0.00005}                                  \\
                                                        & \textcolor[rgb]{0.102,0.11,0.122}{total\_steps =~}\textcolor[rgb]{0.102,0.11,0.122}{20000}                                     \\
                                                        & scheduler =~\textcolor[rgb]{0.102,0.11,0.122}{warmup\_linear}                                                                  \\
                                                        & \textcolor[rgb]{0.212,0.227,0.239}{\textcolor[rgb]{0.102,0.11,0.122}{warmup\_steps =~}\textcolor[rgb]{0.102,0.11,0.122}{250}}  \\ \hline
\multirow{3}{*}{epochs}                                 & \textcolor[rgb]{0.212,0.227,0.239}{\textcolor[rgb]{0.102,0.11,0.122}{max\_epochs =~}\textcolor[rgb]{0.102,0.11,0.122}{0}}      \\
                                                        & \textcolor[rgb]{0.102,0.11,0.122}{max\_steps =~}\textcolor[rgb]{0.102,0.11,0.122}{20000}                                       \\
                                                        & \textcolor[rgb]{0.102,0.11,0.122}{patience =~}\textcolor[rgb]{0.102,0.11,0.122}{1600}                                          \\ \hline
\textcolor[rgb]{0.102,0.11,0.122}{accumulate\_gradient} & 3                                                                                                                              \\ \hline
\multirow{7}{*}{optimizer}                              & \textcolor[rgb]{0.212,0.227,0.239}{\textcolor[rgb]{0.102,0.11,0.122}{AdamW}}                                                   \\
                                                        & \textcolor[rgb]{0.102,0.11,0.122}{beta = 1}\textcolor[rgb]{0.102,0.11,0.122}{0.9}                                              \\
                                                        & \textcolor[rgb]{0.102,0.11,0.122}{beta2 =~}\textcolor[rgb]{0.102,0.11,0.122}{0.999}                                            \\
                                                        & \textcolor[rgb]{0.102,0.11,0.122}{eps =~}\textcolor[rgb]{0.102,0.11,0.122}{1e-8}                                               \\
                                                        & \textcolor[rgb]{0.102,0.11,0.122}{grad\_clip =~}\textcolor[rgb]{0.102,0.11,0.122}{1}                                           \\
                                                        & \textcolor[rgb]{0.102,0.11,0.122}{l2 =~}\textcolor[rgb]{0.102,0.11,0.122}{0.01}                                                \\
                                                        & \textcolor[rgb]{0.102,0.11,0.122}{l2\_is\_weight\_decay =~}\textcolor[rgb]{0.102,0.11,0.122}{true}                             \\ \hline
eval\_frequency                                         & 200                                                                                                                            \\ \hline
dropout                                                 & 0.1                                                                                                                            \\ \bottomrule
\end{tabular}
}
\end{minipage}
\end{table}

\subsection{Word Camouflage NER models}\label{sec:models}

As mentioned above, one of the objectives of this work is to develop a multilingual detector model to address the problem of content evasion by word camouflage from a multilingual perspective. Likewise, another objective is to continue the research carried out in our previous work~\cite{aida_model_2021} focused on the usefulness of semantic similarity as a generalization task at the multilingual level. This is why we include the model developed in the previous work and its baseline, and other remarkable multilingual models. Accordingly, the multilingual models fitted to the Named Entity Recognition task of camouflaged words are presented below. In the following sections of the article, we will refer to the various multilingual models that we have tested using their abbreviations. This is done to make the text more concise and easier to read, while still providing enough information for the reader to understand the context.


\begin{itemize}

    \item \textbf{paraphrase-multilingual-mpnet-base-v2 (MPNET-base)}: 
    Distilled version of the MPNet model from Microsoft~\cite{song2020mpnet} fine-tuned with large-scale paraphrase data using XLM-RoBERTa as the student model. This model is included as a baseline to corroborate the usefulness of the multilingual pre-train in semantic similarity showed in our previous work ~\cite{aida_model_2021}, as it does not includes any fine-tuning on semantic similarity.

    \item \textbf{mstsb-paraphrase-multilingual-mpnet-base-v2 (MPNET-ideal)}:  
    Previous model fitted with multilingual train data from the Semantic Textual Similairty Benchmark (STSb)~\cite{cer-etal-2017-semeval} extended version to 15 languages (mSTSb)~\cite{aida_model_2021}. This model has shown to enhance the performance across languages, outperform monolingual models and the capability of generalize to new tasks. This model has been presented previously in the 22nd International Conference on Intelligent Data Engineering and Automated Learning (IDEAL)~\cite{aida_model_2021}.
    


    \item \textbf{bloomz-560m (BLOOMz)}~\cite{bloomz}: Fine-tuned variant of the pre-trained multilingual BLOOM~\cite{scao2022bloom} and mT5~\cite{mt5} model families on cross-lingual task mixture of 13 training tasks in 46 languages with English prompts capable of following human instructions in dozens of languages zero-shot. The version used is the version of 560M parameters.  

    \item \textbf{xlm-roberta-base (XLM-R)}: Base-sized XLM-RoBERTa~\cite{conneau2020unsupervised} model totalizing $\sim$125M parameters. XLM-RoBERTa is RoBERTa model~\cite{liu2019roberta}, robust version of BERT,  pre-trained on CommonCrawl data containing 100 languages.     
    

    \item \textbf{Model 5 - bert-base-multilingual-cased (mBERT)}: BERT~\cite{devlin2018bert} transformer model pre-trained on a large corpus of 104 languages Wikipedia articles using the self-supervised masked language modelling  (MLM) objective with $\sim$177M parameters.

\end{itemize}

The best multilingual model obtained for the NER task is compared with the monolingual model fine-tuned for the NER task in each language included. Therefore, the following monolingual models are considered as baseline:

\begin{itemize}
    \item \textbf{roberta-base}~\cite{liu2019roberta}: English baseline model. Pre-trained model in English using a masked language modeling (MLM) on Wikipedia articles and BookCorpus~\cite{Zhu_2015_ICCV}.
    
    \item \textbf{roberta-base-bne}~\cite{gutierrezfandino2021spanish}: Spanish baseline model. Masked language model for the Spanish language based on the RoBERTa base model pre-trained using spanish web crawlings  performed by the National Library of Spain from 2009 to 2019.

    \item \textbf{camembert-base}~\cite{martin2020camembert}: French baseline model. 
     Masked language model for French based on the RoBERTa base model pre-trained using the French portion of the Open Super-large Crawled Aggregated coRpus (OSCAR)~\cite{OSCAR}.

    \item \textbf{robit-roberta-base-it}~\cite{gottbert}: Italian baseline model. 
     Masked language model for French based on the RoBERTa base model pre-trained solely on the Italian portion of the OSCAR dataset.     

    \item \textbf{gottbert-base}~\cite{gottbert}: German baseline model. 
     Masked language model for French based on the BERT base model pre-trained solely on the German portion of the OSCAR dataset.

\end{itemize}


The models were fine-tuned using the Spacy interface~\cite{montani_spacy_2020} as the camouflage NER data is in Spacy format.  For the sake of reproducibility, the parameters and hyperparameters used during the training process can be consulted in Table \ref{table:training-params}. A more detailed view of these parameters and training metrics is available at Weight \& Biases\footnote{\href{https://wandb.ai/aida-group/ASOC-LeetSpeakNER-full-XX-MultiNER/overview}{\url{https://wandb.ai/aida-group/ASOC-LeetSpeakNER-full-XX-MultiNER/overview}}}. Additionally, the models are publicly available on Hugging Face\textsuperscript{\ref{foot:model}} either for direct use or for integration into other Spacy pipelines.

\section{Experiments and Results}
\label{sec:results}

The experiments presented in this section aim to develop the best multilingual NER model for word camouflage detection. Then compare this best multilingual model against the monolingual baseline models. Finally, we analyze the detection performance of different camouflage entities by visualizing the confusion matrices of the best multilingual model. As the NER word camouflaged detection task considered in this work consists of four imbalanced mutually exclusive classes (see Subsections~\ref{sec:ner-generator} and \ref{sec:ner-data}), the F1 score metric is reported with its different variants, micro, macro and weighted averages. After all, F1 score variants include both precision and recall because they rely on the model's True Positives (TP), False Positives (FP) and False Negatives (FN). 

The macro-averaged F1 score represents the unweighted mean; this is computing the arithmetic mean of all the per-class F1 scores.
On the other hand, the micro-averaged F1 score computes the proportion of correctly classified observations out of all observations, as it computes a global average F1 score by summing the respective TP, FP, FN values across all classes. Finally, the F1-weighted average is calculated by taking the mean of all per-class F1 scores while considering each class's support.

Since the models are tested on an imbalanced dataset, the F1 macro and F1 weighted averages are preferred. It is important to note that the F1-macro metric allows us to evaluate the models considering that all classes are equally important. At the same time, F1-weighted assigns higher contributions to the classes with more examples in the dataset. 
As explained in the Methodology Section \ref{sec:ner-data} not all types of camouflage are applied in the same proportion, so we have paid particular attention to the F1-weighted mean. Similarly, note that test results are reported in general and broken down by dataset. The overall result is obtained by considering all instances of the datasets as a whole and not the average of the individual results.


\subsection{Multilingual NER word camouflage models}

\begin{table}
\centering
\caption{F1-Macro, F1-micro and F1 weighted average test results for the multilingual models according to the overall and each dataset, following the nomenclature and order shown in Subsection \ref{sec:models}. In bold the best result, in italics the second best.}
\label{table:XX-test-results}
\begin{tblr}{
  width = \linewidth,
  colspec = {Q[158]Q[181]Q[117]Q[117]Q[117]Q[117]Q[117]},
  cells = {c},
  cell{2}{1} = {r=3}{},
  cell{5}{1} = {r=3}{},
  cell{8}{1} = {r=3}{},
  cell{11}{1} = {r=3}{},
  cell{14}{1} = {r=3}{},
  hline{2} = {-}{},
  hline{5,8,11,14} = {2-7}{},
  hline{17} = {-}{},
}
                   &             & MPNET-base         & MPNET-ideal         & BLOOMz & XLM-R         & mBERT     \\
{News~\\Comentary} & F1-Macro    & 0.9398          & 0.9455          & 0.8221  & \textbf{0.9470} & \textit{0.9458} \\
                   & F1-Micro    & 0.9892          & \textit{0.9908} & 0.9653  & \textbf{0.9916} & 0.9900          \\
                   & F1-Weighted & 0.8855          & \textbf{0.9019} & 0.6845  & 0.8843          & 0.8915          \\
ParaCrawl          & F1-Macro    & 0.9306          & \textit{0.9316} & 0.7942  & \textbf{0.9336} & 0.9274          \\
                   & F1-Micro    & 0.9880          & \textit{0.9887} & 0.9617  & \textbf{0.9890} & 0.9846          \\
                   & F1-Weighted & 0.8720          & \textbf{0.8760} & 0.6457  & 0.8633          & 0.8643          \\
TED2020            & F1-Macro    & 0.9400          & \textbf{0.9437} & 0.8062  & 0.9397          & \textit{0.9404} \\
                   & F1-Micro    & 0.9880          & \textbf{0.9893} & 0.9597  & \textit{0.9883} & 0.9867          \\
                   & F1-Weighted & 0.8863          & \textbf{0.8933} & 0.6652  & 0.8746          & \textit{0.8837} \\
WikiMatrix         & F1-Macro    & 0.9195          & \textbf{0.9291} & 0.7860  & 0.9239          & \textit{0.9279} \\
                   & F1-Micro    & 0.9810          & \textbf{0.9839} & 0.9465  & \textit{0.9827} & 0.9816          \\
                   & F1-Weighted & 0.8516          & \textbf{0.8664} & 0.6324  & 0.8427          & \textit{0.8598} \\
Overall            & F1-Macro    & 0.9308          & \textbf{0.9350} & 0.7971  & \textit{0.9334} & 0.9321          \\
                   & F1-Micro    & 0.9866          & \textbf{0.9880} & 0.9582  & \textit{0.9876} & 0.9849          \\
                   & F1-Weighted & \textit{0.8712} & \textbf{0.8795} & 0.6499  & 0.8620          & 0.8698          
\end{tblr}
\end{table}

\begin{table}
\centering
\caption{F1-Macro, F1-micro, and F1 weighted average test results for the best multilingual model (i.e., MPNET-ideal from Subsection \ref{sec:models}) and each monolingual baseline model for the languages considered according to the overall and each data set.}
\label{table:mono-vs-multi}
\resizebox{\linewidth}{!}{%
\begin{tabular}{cc||cc||cl||cc||cc||cc}
\vcell{}                                                                  & \vcell{}            & \multicolumn{2}{c||}{\vcell{\textbf{EN}}}                                                        & \multicolumn{2}{c||}{\vcell{\textbf{ES}}}                                                        & \multicolumn{2}{c||}{\vcell{\textbf{FR}}}                                                        & \multicolumn{2}{c||}{\vcell{\textbf{IT}}}                                                        & \multicolumn{2}{c}{\vcell{\textbf{DE}}}                                                           \\[-\rowheight]
\printcelltop                                                             & \printcelltop       & \multicolumn{2}{c||}{\printcelltop}                                                              & \multicolumn{2}{c||}{\printcelltop}                                                              & \multicolumn{2}{c||}{\printcellmiddle}                                                           & \multicolumn{2}{c||}{\printcellmiddle}                                                           & \multicolumn{2}{c}{\printcellmiddle}                                                              \\ 
\toprule
\multirow{6}{*}{\begin{tabular}[c]{@{}c@{}}News~\\Comentary\end{tabular}} & \vcell{F1-Macro}    & \multirow{3}{*}{\begin{tabular}[c]{@{}c@{}}EN~\\Model\end{tabular}}    & \vcell{0.9172}          & \multirow{3}{*}{\begin{tabular}[c]{@{}c@{}}ES~\\Model\end{tabular}}    & \vcell{0.9501}          & \multirow{3}{*}{\begin{tabular}[c]{@{}c@{}}FR~\\Model\end{tabular}}    & \vcell{0.9365}          & \multirow{3}{*}{\begin{tabular}[c]{@{}c@{}}IT~\\Model\end{tabular}}    & \vcell{0.8270}          & \multirow{3}{*}{\begin{tabular}[c]{@{}c@{}}DE~\\Model\end{tabular}}    & \vcell{0.9607}           \\[-\rowheight]
                                                                          & \printcelltop       &                                                                        & \printcelltop           &                                                                        & \printcellmiddle        &                                                                        & \printcellmiddle        &                                                                        & \printcellmiddle        &                                                                        & \printcellmiddle         \\
                                                                          & \vcell{F1-Micro}    &                                                                        & \vcell{0.9864}          &                                                                        & \vcell{0.9901}          &                                                                        & \vcell{0.9880}          &                                                                        & \vcell{0.9612}          &                                                                        & \vcell{0.9909}           \\[-\rowheight]
                                                                          & \printcelltop       &                                                                        & \printcelltop           &                                                                        & \printcellmiddle        &                                                                        & \printcellmiddle        &                                                                        & \printcellmiddle        &                                                                        & \printcellmiddle         \\
                                                                          & \vcell{F1-Weighted} &                                                                        & \vcell{\textbf{0.8522}} &                                                                        & \vcell{0.9025}          &                                                                        & \vcell{\textbf{0.8705}} &                                                                        & \vcell{0.6982}          &                                                                        & \vcell{0.9210}           \\[-\rowheight]
                                                                          & \printcelltop       &                                                                        & \printcelltop           &                                                                        & \printcellmiddle        &                                                                        & \printcellmiddle        &                                                                        & \printcellmiddle        &                                                                        & \printcellmiddle         \\ 
\cline{2-12}
                                                                          & \vcell{F1-Macro}    & \multirow{3}{*}{\begin{tabular}[c]{@{}c@{}}MPNET-\\ideal\end{tabular}} & \vcell{0.9167}          & \multirow{3}{*}{\begin{tabular}[c]{@{}c@{}}MPNET-\\ideal\end{tabular}} & \vcell{0.9487}          & \multirow{3}{*}{\begin{tabular}[c]{@{}c@{}}MPNET-\\ideal\end{tabular}} & \vcell{0.8771}          & \multirow{3}{*}{\begin{tabular}[c]{@{}c@{}}MPNET-\\ideal\end{tabular}} & \vcell{0.8783}          & \multirow{3}{*}{\begin{tabular}[c]{@{}c@{}}MPNET-\\ideal\end{tabular}} & \vcell{0.9733}           \\[-\rowheight]
                                                                          & \printcelltop       &                                                                        & \printcelltop           &                                                                        & \printcellmiddle        &                                                                        & \printcellmiddle        &                                                                        & \printcellmiddle        &                                                                        & \printcellmiddle         \\
                                                                          & \vcell{F1-Micro}    &                                                                        & \vcell{0.9868}          &                                                                        & \vcell{0.9917}          &                                                                        & \vcell{0.9756}          &                                                                        & \vcell{0.9835}          &                                                                        & \vcell{0.9940}           \\[-\rowheight]
                                                                          & \printcelltop       &                                                                        & \printcelltop           &                                                                        & \printcellmiddle        &                                                                        & \printcellmiddle        &                                                                        & \printcellmiddle        &                                                                        & \printcellmiddle         \\
                                                                          & \vcell{F1-Weighted} &                                                                        & \vcell{0.8515}          &                                                                        & \vcell{\textbf{0.9081}} &                                                                        & \vcell{0.7872}          &                                                                        & \vcell{\textbf{0.8133}} &                                                                        & \vcell{\textbf{0.9484}}  \\[-\rowheight]
                                                                          & \printcelltop       &                                                                        & \printcelltop           &                                                                        & \printcellmiddle        &                                                                        & \printcellmiddle        &                                                                        & \printcellmiddle        &                                                                        & \printcellmiddle         \\ 
\toprule
\multirow{6}{*}{ParaCrawl}                                                & \vcell{F1-Macro}    & \multirow{3}{*}{\begin{tabular}[c]{@{}c@{}}EN~\\Model\end{tabular}}    & \vcell{0.8381}          & \multirow{3}{*}{\begin{tabular}[c]{@{}c@{}}ES~\\Model\end{tabular}}    & \vcell{0.9387}          & \multirow{3}{*}{\begin{tabular}[c]{@{}c@{}}FR~\\Model\end{tabular}}    & \vcell{0.9338}          & \multirow{3}{*}{\begin{tabular}[c]{@{}c@{}}IT~\\Model\end{tabular}}    & \vcell{0.8265}          & \multirow{3}{*}{\begin{tabular}[c]{@{}c@{}}DE~\\Model\end{tabular}}    & \vcell{0.9521}           \\[-\rowheight]
                                                                          & \printcelltop       &                                                                        & \printcelltop           &                                                                        & \printcellmiddle        &                                                                        & \printcellmiddle        &                                                                        & \printcellmiddle        &                                                                        & \printcellmiddle         \\
                                                                          & \vcell{F1-Micro}    &                                                                        & \vcell{0.9639}          &                                                                        & \vcell{0.9852}          &                                                                        & \vcell{0.9863}          &                                                                        & \vcell{0.9586}          &                                                                        & \vcell{0.9884}           \\[-\rowheight]
                                                                          & \printcelltop       &                                                                        & \printcelltop           &                                                                        & \printcellmiddle        &                                                                        & \printcellmiddle        &                                                                        & \printcellmiddle        &                                                                        & \printcellmiddle         \\
                                                                          & \vcell{F1-Weighted} &                                                                        & \vcell{0.7376}          &                                                                        & \vcell{0.8921}          &                                                                        & \vcell{0.8713}          &                                                                        & \vcell{0.6915}          &                                                                        & \vcell{0.9048}           \\[-\rowheight]
                                                                          & \printcelltop       &                                                                        & \printcelltop           &                                                                        & \printcellmiddle        &                                                                        & \printcellmiddle        &                                                                        & \printcellmiddle        &                                                                        & \printcellmiddle         \\ 
\cline{2-12}
                                                                          & \vcell{F1-Macro}    & \multirow{3}{*}{\begin{tabular}[c]{@{}c@{}}MPNET-\\ideal\end{tabular}} & \vcell{0.8690}          & \multirow{3}{*}{\begin{tabular}[c]{@{}c@{}}MPNET-\\ideal\end{tabular}} & \vcell{0.9422}          & \multirow{3}{*}{\begin{tabular}[c]{@{}c@{}}MPNET-\\ideal\end{tabular}} & \vcell{0.9373}          & \multirow{3}{*}{\begin{tabular}[c]{@{}c@{}}MPNET-\\ideal\end{tabular}} & \vcell{0.9396}          & \multirow{3}{*}{\begin{tabular}[c]{@{}c@{}}MPNET-\\ideal\end{tabular}} & \vcell{0.9504}           \\[-\rowheight]
                                                                          & \printcelltop       &                                                                        & \printcelltop           &                                                                        & \printcellmiddle        &                                                                        & \printcellmiddle        &                                                                        & \printcellmiddle        &                                                                        & \printcellmiddle         \\
                                                                          & \vcell{F1-Micro}    &                                                                        & \vcell{0.9809}          &                                                                        & \vcell{0.9903}          &                                                                        & \vcell{0.9909}          &                                                                        & \vcell{0.9897}          &                                                                        & \vcell{0.9909}           \\[-\rowheight]
                                                                          & \printcelltop       &                                                                        & \printcelltop           &                                                                        & \printcellmiddle        &                                                                        & \printcellmiddle        &                                                                        & \printcellmiddle        &                                                                        & \printcellmiddle         \\
                                                                          & \vcell{F1-Weighted} &                                                                        & \vcell{\textbf{0.7750}} &                                                                        & \vcell{\textbf{0.8977}} &                                                                        & \vcell{\textbf{0.8847}} &                                                                        & \vcell{\textbf{0.8849}} &                                                                        & \vcell{\textbf{0.9126}}  \\[-\rowheight]
                                                                          & \printcelltop       &                                                                        & \printcelltop           &                                                                        & \printcellmiddle        &                                                                        & \printcellmiddle        &                                                                        & \printcellmiddle        &                                                                        & \printcellmiddle         \\ 
\toprule
\multirow{6}{*}{TED2020}                                                  & \vcell{F1-Macro}    & \multirow{3}{*}{\begin{tabular}[c]{@{}c@{}}EN~\\Model\end{tabular}}    & \vcell{0.9102}          & \multirow{3}{*}{\begin{tabular}[c]{@{}c@{}}ES~\\Model\end{tabular}}    & \vcell{0.9552}          & \multirow{3}{*}{\begin{tabular}[c]{@{}c@{}}FR~\\Model\end{tabular}}    & \vcell{0.9325}          & \multirow{3}{*}{\begin{tabular}[c]{@{}c@{}}IT~\\Model\end{tabular}}    & \vcell{0.8535}          & \multirow{3}{*}{\begin{tabular}[c]{@{}c@{}}DE~\\Model\end{tabular}}    & \vcell{0.9577}           \\[-\rowheight]
                                                                          & \printcelltop       &                                                                        & \printcelltop           &                                                                        & \printcellmiddle        &                                                                        & \printcellmiddle        &                                                                        & \printcellmiddle        &                                                                        & \printcellmiddle         \\
                                                                          & \vcell{F1-Micro}    &                                                                        & \vcell{0.9806}          &                                                                        & \vcell{0.9890}          &                                                                        & \vcell{0.9840}          &                                                                        & \vcell{0.9612}          &                                                                        & \vcell{0.9882}           \\[-\rowheight]
                                                                          & \printcelltop       &                                                                        & \printcelltop           &                                                                        & \printcellmiddle        &                                                                        & \printcellmiddle        &                                                                        & \printcellmiddle        &                                                                        & \printcellmiddle         \\
                                                                          & \vcell{F1-Weighted} &                                                                        & \vcell{0.8286}          &                                                                        & \vcell{0.9104}          &                                                                        & \vcell{0.8704}          &                                                                        & \vcell{0.7353}          &                                                                        & \vcell{0.9140}           \\[-\rowheight]
                                                                          & \printcelltop       &                                                                        & \printcelltop           &                                                                        & \printcellmiddle        &                                                                        & \printcellmiddle        &                                                                        & \printcellmiddle        &                                                                        & \printcellmiddle         \\ 
\cline{2-12}
                                                                          & \vcell{F1-Macro}    & \multirow{3}{*}{\begin{tabular}[c]{@{}c@{}}MPNET-\\ideal\end{tabular}} & \vcell{0.9224}          & \multirow{3}{*}{\begin{tabular}[c]{@{}c@{}}MPNET-\\ideal\end{tabular}} & \vcell{0.9528}          & \multirow{3}{*}{\begin{tabular}[c]{@{}c@{}}MPNET-\\ideal\end{tabular}} & \vcell{0.9376}          & \multirow{3}{*}{\begin{tabular}[c]{@{}c@{}}MPNET-\\ideal\end{tabular}} & \vcell{0.9469}          & \multirow{3}{*}{\begin{tabular}[c]{@{}c@{}}MPNET-\\ideal\end{tabular}} & \vcell{0.9574}           \\[-\rowheight]
                                                                          & \printcelltop       &                                                                        & \printcelltop           &                                                                        & \printcellmiddle        &                                                                        & \printcellmiddle        &                                                                        & \printcellmiddle        &                                                                        & \printcellmiddle         \\
                                                                          & \vcell{F1-Micro}    &                                                                        & \vcell{0.9854}          &                                                                        & \vcell{0.9909}          &                                                                        & \vcell{0.9892}          &                                                                        & \vcell{0.9898}          &                                                                        & \vcell{0.9911}           \\[-\rowheight]
                                                                          & \printcelltop       &                                                                        & \printcelltop           &                                                                        & \printcellmiddle        &                                                                        & \printcellmiddle        &                                                                        & \printcellmiddle        &                                                                        & \printcellmiddle         \\
                                                                          & \vcell{F1-Weighted} &                                                                        & \vcell{\textbf{0.8500}} &                                                                        & \vcell{\textbf{0.9153}} &                                                                        & \vcell{\textbf{0.8844}} &                                                                        & \vcell{\textbf{0.8972}} &                                                                        & \vcell{\textbf{0.9181}}  \\[-\rowheight]
                                                                          & \printcelltop       &                                                                        & \printcelltop           &                                                                        & \printcellmiddle        &                                                                        & \printcellmiddle        &                                                                        & \printcellmiddle        &                                                                        & \printcellmiddle         \\ 
\toprule
\multirow{6}{*}{WikiMatrix}                                               & \vcell{F1-Macro}    & \multirow{3}{*}{\begin{tabular}[c]{@{}c@{}}EN~\\Model\end{tabular}}    & \vcell{0.8889}          & \multirow{3}{*}{\begin{tabular}[c]{@{}c@{}}ES~\\Model\end{tabular}}    & \vcell{0.9323}          & \multirow{3}{*}{\begin{tabular}[c]{@{}c@{}}FR~\\Model\end{tabular}}    & \vcell{0.9024}          & \multirow{3}{*}{\begin{tabular}[c]{@{}c@{}}IT~\\Model\end{tabular}}    & \vcell{0.8332}          & \multirow{3}{*}{\begin{tabular}[c]{@{}c@{}}DE~\\Model\end{tabular}}    & \vcell{0.9417}           \\[-\rowheight]
                                                                          & \printcelltop       &                                                                        & \printcelltop           &                                                                        & \printcellmiddle        &                                                                        & \printcellmiddle        &                                                                        & \printcellmiddle        &                                                                        & \printcellmiddle         \\
                                                                          & \vcell{F1-Micro}    &                                                                        & \vcell{0.9724}          &                                                                        & \vcell{0.9805}          &                                                                        & \vcell{0.9751}          &                                                                        & \vcell{0.9524}          &                                                                        & \vcell{0.9813}           \\[-\rowheight]
                                                                          & \printcelltop       &                                                                        & \printcelltop           &                                                                        & \printcellmiddle        &                                                                        & \printcellmiddle        &                                                                        & \printcellmiddle        &                                                                        & \printcellmiddle         \\
                                                                          & \vcell{F1-Weighted} &                                                                        & \vcell{0.7998}          &                                                                        & \vcell{0.8720}          &                                                                        & \vcell{0.8186}          &                                                                        & \vcell{0.6991}          &                                                                        & \vcell{\textbf{0.8829}}  \\[-\rowheight]
                                                                          & \printcelltop       &                                                                        & \printcelltop           &                                                                        & \printcellmiddle        &                                                                        & \printcellmiddle        &                                                                        & \printcellmiddle        &                                                                        & \printcellmiddle         \\ 
\cline{2-12}
                                                                          & \vcell{F1-Macro}    & \multirow{3}{*}{\begin{tabular}[c]{@{}c@{}}MPNET-\\ideal\end{tabular}} & \vcell{0.9032}          & \multirow{3}{*}{\begin{tabular}[c]{@{}c@{}}MPNET-\\ideal\end{tabular}} & \vcell{0.9349}          & \multirow{3}{*}{\begin{tabular}[c]{@{}c@{}}MPNET-\\ideal\end{tabular}} & \vcell{0.9198}          & \multirow{3}{*}{\begin{tabular}[c]{@{}c@{}}MPNET-\\ideal\end{tabular}} & \vcell{0.9475}          & \multirow{3}{*}{\begin{tabular}[c]{@{}c@{}}MPNET-\\ideal\end{tabular}} & \vcell{0.9380}           \\[-\rowheight]
                                                                          & \printcelltop       &                                                                        & \printcelltop           &                                                                        & \printcellmiddle        &                                                                        & \printcellmiddle        &                                                                        & \printcellmiddle        &                                                                        & \printcellmiddle         \\
                                                                          & \vcell{F1-Micro}    &                                                                        & \vcell{0.9785}          &                                                                        & \vcell{0.9849}          &                                                                        & \vcell{0.9831}          &                                                                        & \vcell{0.9882}          &                                                                        & \vcell{0.9852}           \\[-\rowheight]
                                                                          & \printcelltop       &                                                                        & \printcelltop           &                                                                        & \printcellmiddle        &                                                                        & \printcellmiddle        &                                                                        & \printcellmiddle        &                                                                        & \printcellmiddle         \\
                                                                          & \vcell{F1-Weighted} &                                                                        & \vcell{\textbf{0.8245}} &                                                                        & \vcell{\textbf{0.8805}} &                                                                        & \vcell{\textbf{0.8503}} &                                                                        & \vcell{\textbf{0.8954}} &                                                                        & \vcell{0.8827}           \\[-\rowheight]
                                                                          & \printcelltop       &                                                                        & \printcelltop           &                                                                        & \printcellmiddle        &                                                                        & \printcellmiddle        &                                                                        & \printcellmiddle        &                                                                        & \printcellmiddle         \\ 
\toprule
\multirow{6}{*}{Overall}                                                  & \vcell{F1-Macro}    & \multirow{3}{*}{\begin{tabular}[c]{@{}c@{}}EN~\\Model\end{tabular}}    & \vcell{0.8749}          & \multirow{3}{*}{\begin{tabular}[c]{@{}c@{}}ES~\\Model\end{tabular}}    & \vcell{0.9434}          & \multirow{3}{*}{\begin{tabular}[c]{@{}c@{}}FR~\\Model\end{tabular}}    & \vcell{0.9254}          & \multirow{3}{*}{\begin{tabular}[c]{@{}c@{}}IT~\\Model\end{tabular}}    & \vcell{0.8364}          & \multirow{3}{*}{\begin{tabular}[c]{@{}c@{}}DE~\\Model\end{tabular}}    & \vcell{0.9512}           \\[-\rowheight]
                                                                          & \printcelltop       &                                                                        & \printcelltop           &                                                                        & \printcellmiddle        &                                                                        & \printcellmiddle        &                                                                        & \printcellmiddle        &                                                                        & \printcellmiddle         \\
                                                                          & \vcell{F1-Micro}    &                                                                        & \vcell{0.9712}          &                                                                        & \vcell{0.9863}          &                                                                        & \vcell{0.9834}          &                                                                        & \vcell{0.9578}          &                                                                        & \vcell{0.9869}           \\[-\rowheight]
                                                                          & \printcelltop       &                                                                        & \printcelltop           &                                                                        & \printcellmiddle        &                                                                        & \printcellmiddle        &                                                                        & \printcellmiddle        &                                                                        & \printcellmiddle         \\
                                                                          & \vcell{F1-Weighted} &                                                                        & \vcell{0.7831}          &                                                                        & \vcell{0.8938}          &                                                                        & \vcell{0.8572}          &                                                                        & \vcell{0.7061}          &                                                                        & \vcell{0.9020}           \\[-\rowheight]
                                                                          & \printcelltop       &                                                                        & \printcelltop           &                                                                        & \printcellmiddle        &                                                                        & \printcellmiddle        &                                                                        & \printcellmiddle        &                                                                        & \printcellmiddle         \\ 
\cline{2-12}
                                                                          & \vcell{F1-Macro}    & \multirow{3}{*}{\begin{tabular}[c]{@{}c@{}}MPNET-\\ideal\end{tabular}} & \vcell{0.8958}          & \multirow{3}{*}{\begin{tabular}[c]{@{}c@{}}MPNET-\\ideal\end{tabular}} & \vcell{0.9445}          & \multirow{3}{*}{\begin{tabular}[c]{@{}c@{}}MPNET-\\ideal\end{tabular}} & \vcell{0.9320}          & \multirow{3}{*}{\begin{tabular}[c]{@{}c@{}}MPNET-\\ideal\end{tabular}} & \vcell{0.9439}          & \multirow{3}{*}{\begin{tabular}[c]{@{}c@{}}MPNET-\\ideal\end{tabular}} & \vcell{0.9495}           \\[-\rowheight]
                                                                          & \printcelltop       &                                                                        & \printcelltop           &                                                                        & \printcellmiddle        &                                                                        & \printcellmiddle        &                                                                        & \printcellmiddle        &                                                                        & \printcellmiddle         \\
                                                                          & \vcell{F1-Micro}    &                                                                        & \vcell{0.9817}          &                                                                        & \vcell{0.9898}          &                                                                        & \vcell{0.9886}          &                                                                        & \vcell{0.9893}          &                                                                        & \vcell{0.9898}           \\[-\rowheight]
                                                                          & \printcelltop       &                                                                        & \printcelltop           &                                                                        & \printcellmiddle        &                                                                        & \printcellmiddle        &                                                                        & \printcellmiddle        &                                                                        & \printcellmiddle         \\
                                                                          & \vcell{F1-Weighted} &                                                                        & \vcell{\textbf{0.8126}} &                                                                        & \vcell{\textbf{0.9002}} &                                                                        & \vcell{\textbf{0.8739}} &                                                                        & \vcell{\textbf{0.8913}} &                                                                        & \vcell{\textbf{0.9069}}  \\[-\rowheight]
                                                                          & \printcelltop       &                                                                        & \printcelltop           &                                                                        & \printcellmiddle        &                                                                        & \printcellmiddle        &                                                                        & \printcellmiddle        &                                                                        & \printcellmiddle         \\
\hline
\end{tabular}
}
\end{table}

The results of NER word camouflage detection in the test partitions for the different trained multilingual models are presented in Table~\ref{table:XX-test-results}.

From these results, we can see how MPNET-ideal (see Section \ref{sec:models}), the one previously developed using the semantic textual similarity pre-training task with the multilingual extended mSTSb dataset~\cite{aida_model_2021}, shows the best performance in most datasets and is the best in general. 
 It should be noted that MPNET-ideal outperforms MPNET-base, which corresponds to the same model and architecture, but without pre-training in mSTSb. This result corroborates the suitability of the semantic similarity task as a method of providing generality knowledge to a model and the benefit of the multilingual extension of a dataset shown in our previous work~\cite{aida_model_2021}.

Remarkably, the models perform adequately in the different scenarios, since there are no significant differences in the results obtained between the other datasets. However, the WikiMatrix and ParaCrawl datasets show the lowest scores. This could be due to the diversity of symbols present in the data, since both are the result of multilingual Wikipedia and Internet crawls, which can differ between languages and also include URLs, programming code, or mathematical formulas with the variability of symbols that this implies. On the contrary, the News Commentary and TED2020 datasets show better scores since they correspond to natural text in a formal and informal style, respectively, but with less variability of symbols. It should also be noted that good results are obtained in a few shot scenario, such as the News Commentary dataset, which is the dataset with the lowest number of instances (see Table~\ref{table:data-breakdown}). This indicates generalization and transfer knowledge capabilities to different scenarios.

\subsection{Monolingual Baseline NER word camouflage models}

The best multilingual model, MPNET-ideal, is compared with the monolingual baseline models. The results of these models and their comparison are shown in Table~\ref{table:mono-vs-multi}. 

In particular, the test results in the overall dataset show how the multilingual model outperforms the monolingual baseline models. 
The most significant difference is found in the Italian language. The Italian monolingual baseline model has the lowest weighted F1 score of 0.7061 across languages; however, the multilingual model improves the score to 0.8913. Similarly, in the case of English and French, the baseline performance with an F1 score weighted of 0.7831 and 0.8572 is improved to 0.8126 and 0.8739, respectively. Although not to the same extent, the Spanish and German languages also improve the baseline model scores.

Our experiments are consistent with previous result~\cite{aida_model_2021} as corroborate the usefulness of extending a dataset at the multilingual level to improve the performance of a multilingual model over those of monolingual models. This is an advantage in terms of using multilingual models over monolingual ones. Lower computational cost is required, and feasibility and applicability are increased since instead of a monolingual model to detect camouflage in each language, a single model can be used for all of them with better and more consistent results across languages.

\subsection{Multilingual NER Confusion Matrix Analisys}

\begin{figure}[tpb]
\centering
\begin{subfigure}[h]{0.54\linewidth}
\includegraphics[width=\linewidth]{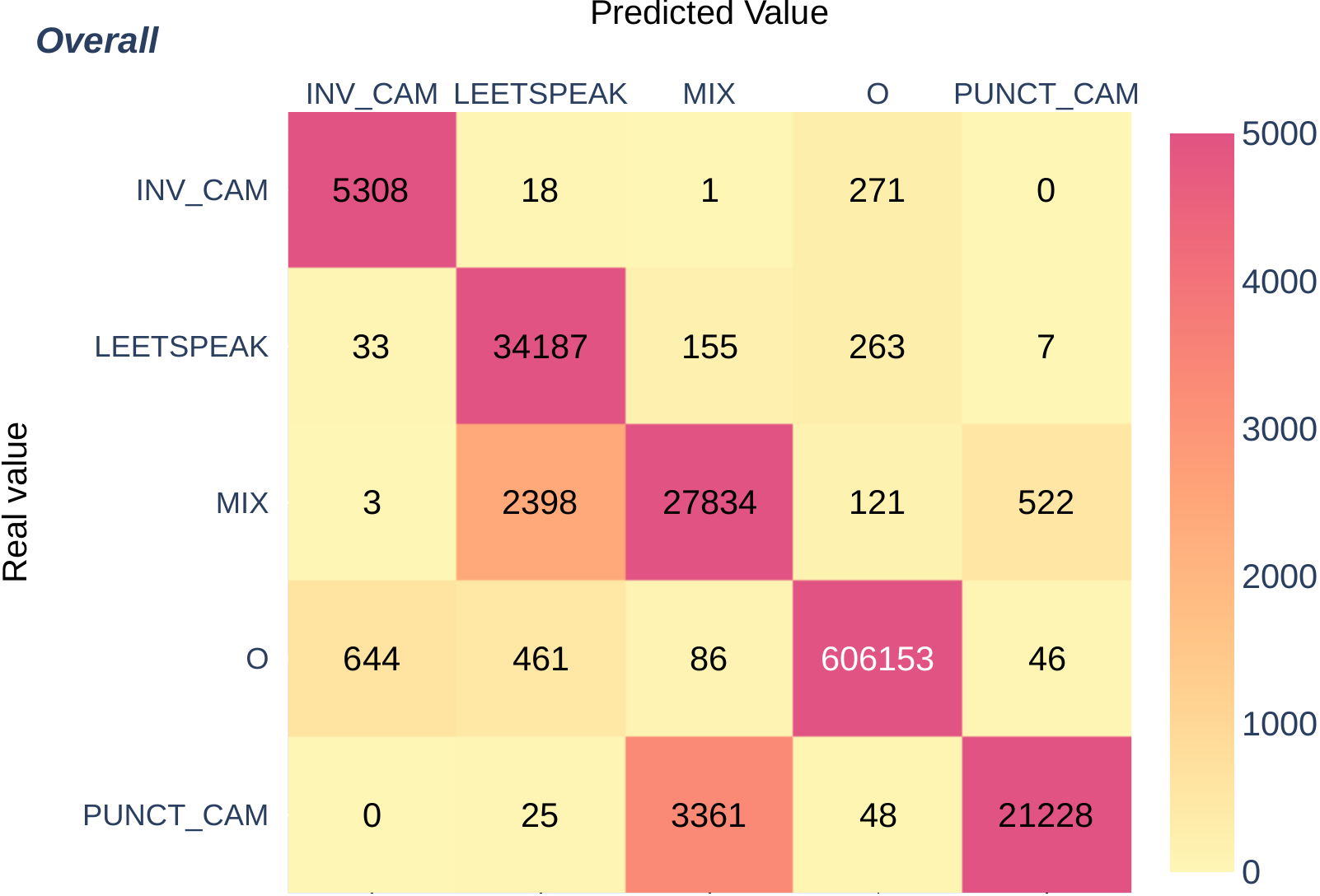}
\caption{}
\label{cm-EN}
\end{subfigure}

\vspace{0.3cm}
\begin{subfigure}[h]{0.47\linewidth}
\includegraphics[width=\linewidth]{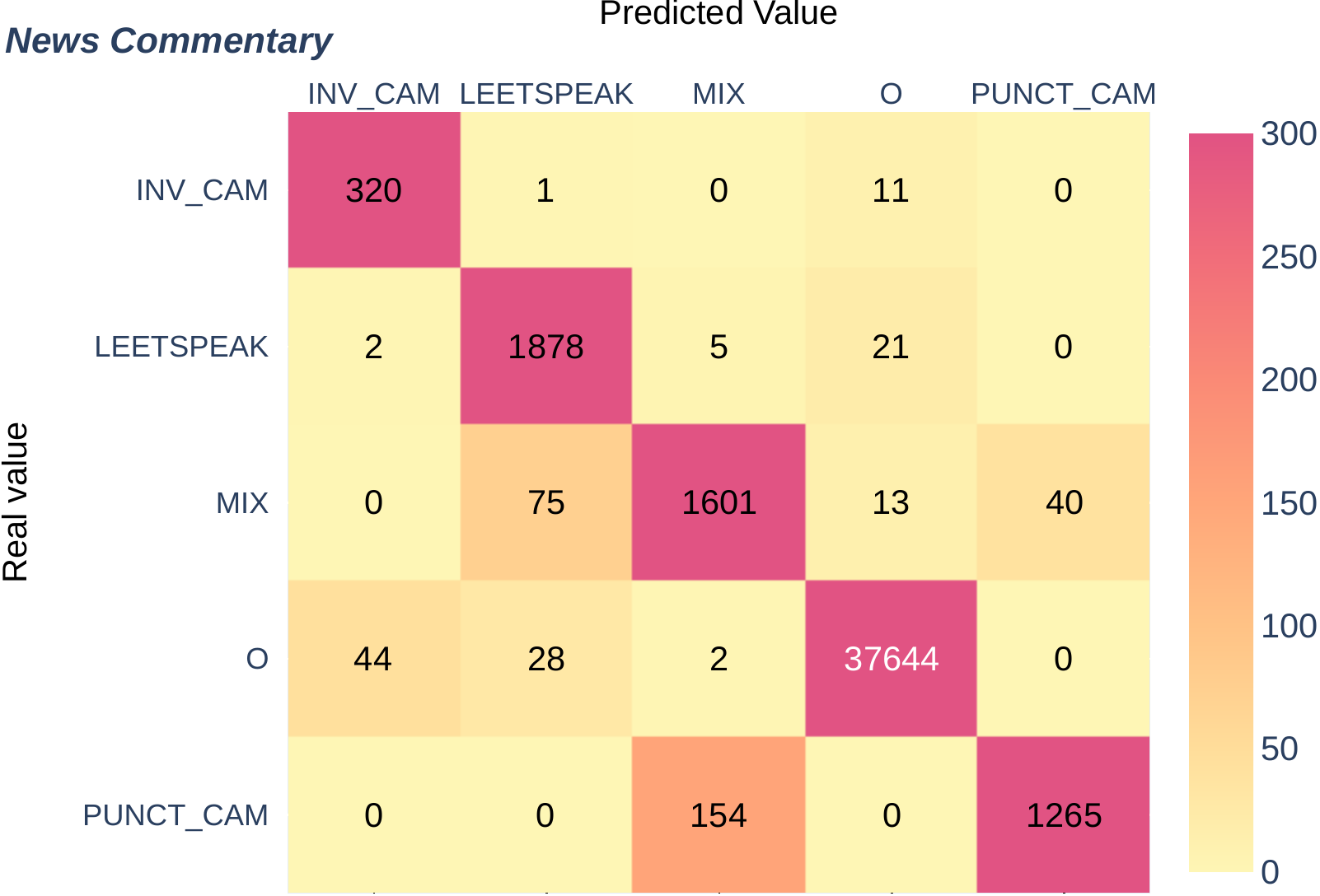}
\caption{}
\label{cm-News_Commentary}
\end{subfigure}
\hspace{0.5cm}
\begin{subfigure}[h]{0.47\linewidth}
\includegraphics[width=\linewidth]{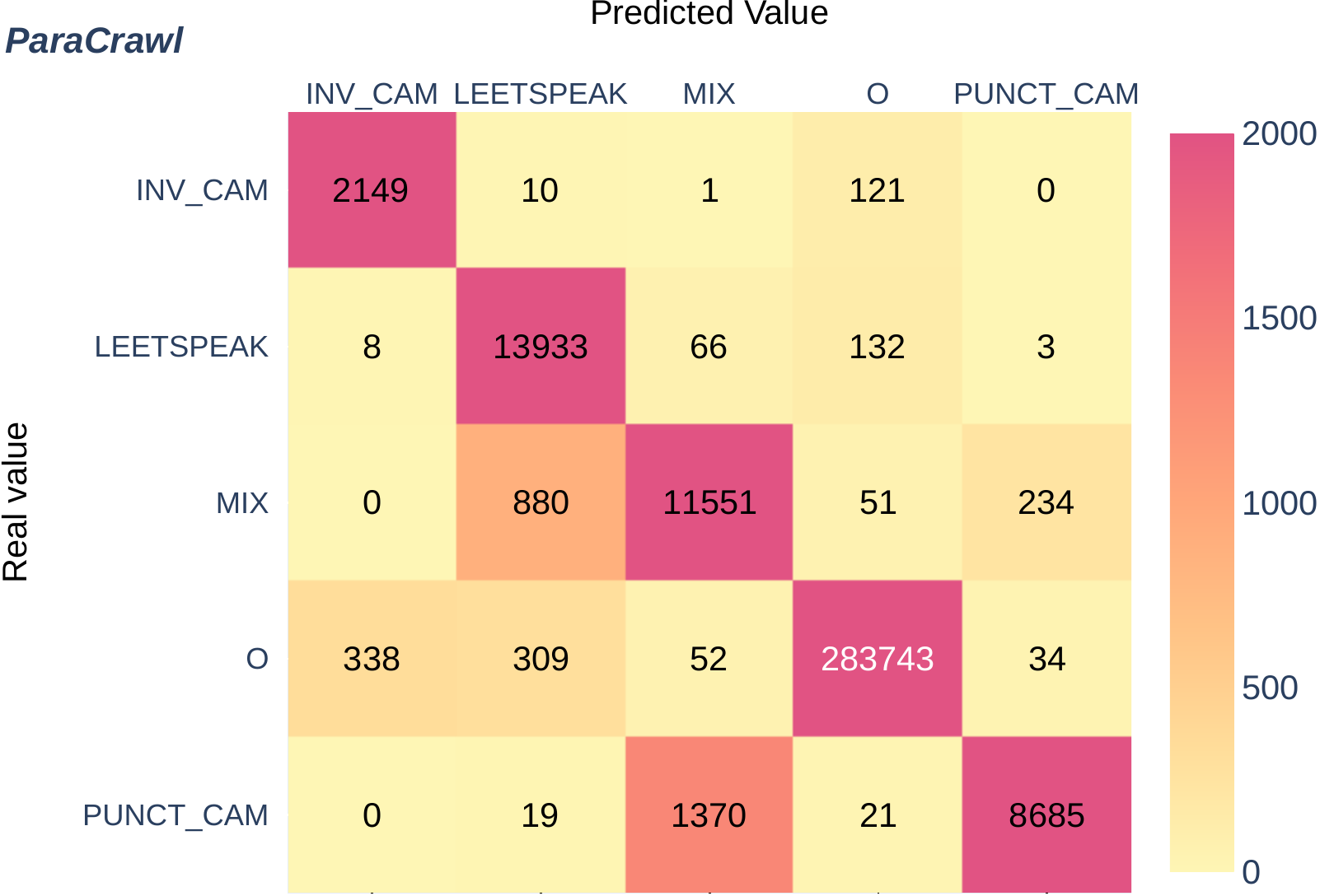}
\caption{}
\label{cm-ParaCrawl}
\end{subfigure}

\vspace{0.3cm}

\begin{subfigure}[h]{0.47\linewidth}
\includegraphics[width=\linewidth]{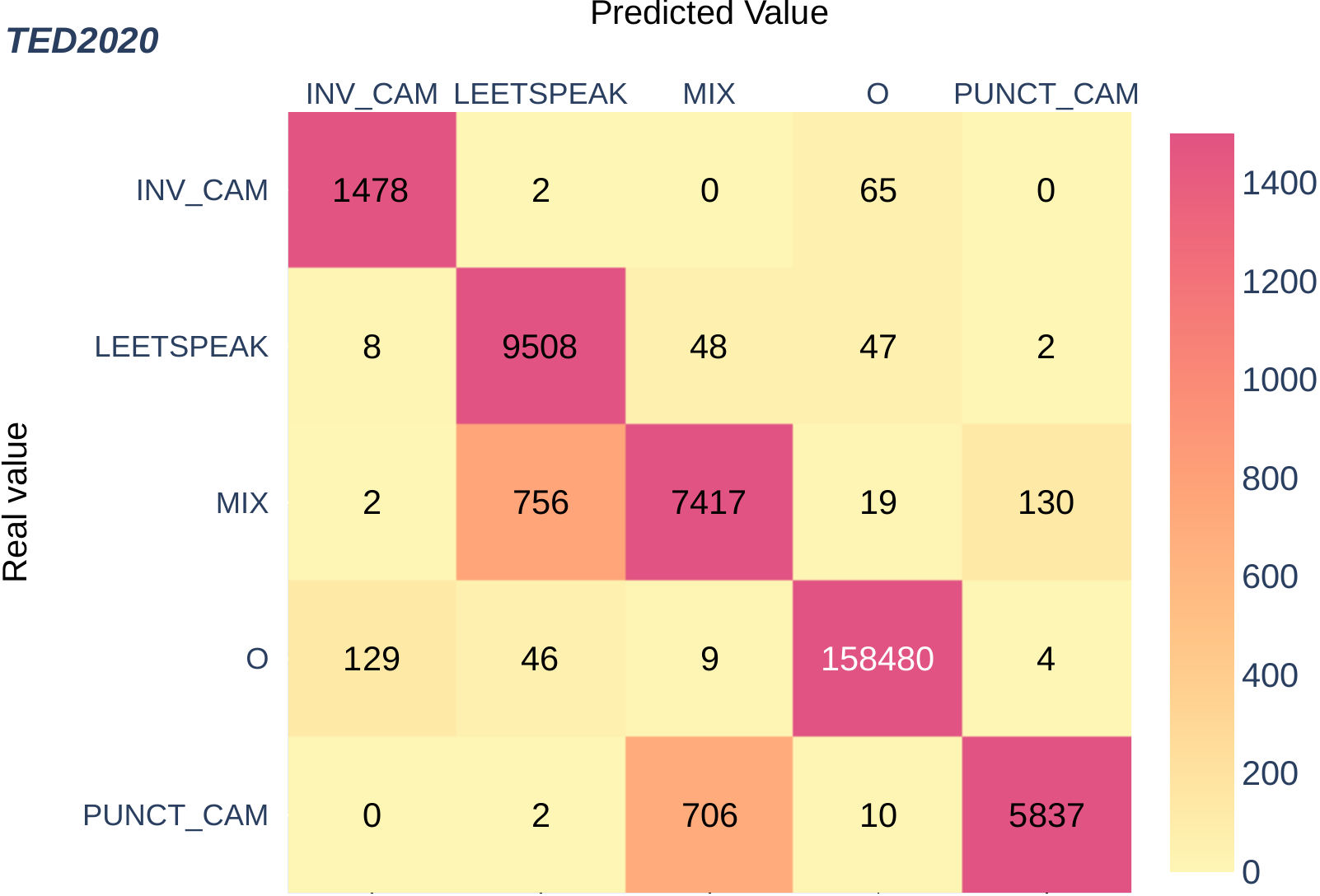}
\caption{}
\label{cm-TED2020}
\end{subfigure}
\hspace{0.5cm}
\begin{subfigure}[h]{0.47\linewidth}
\includegraphics[width=\linewidth]{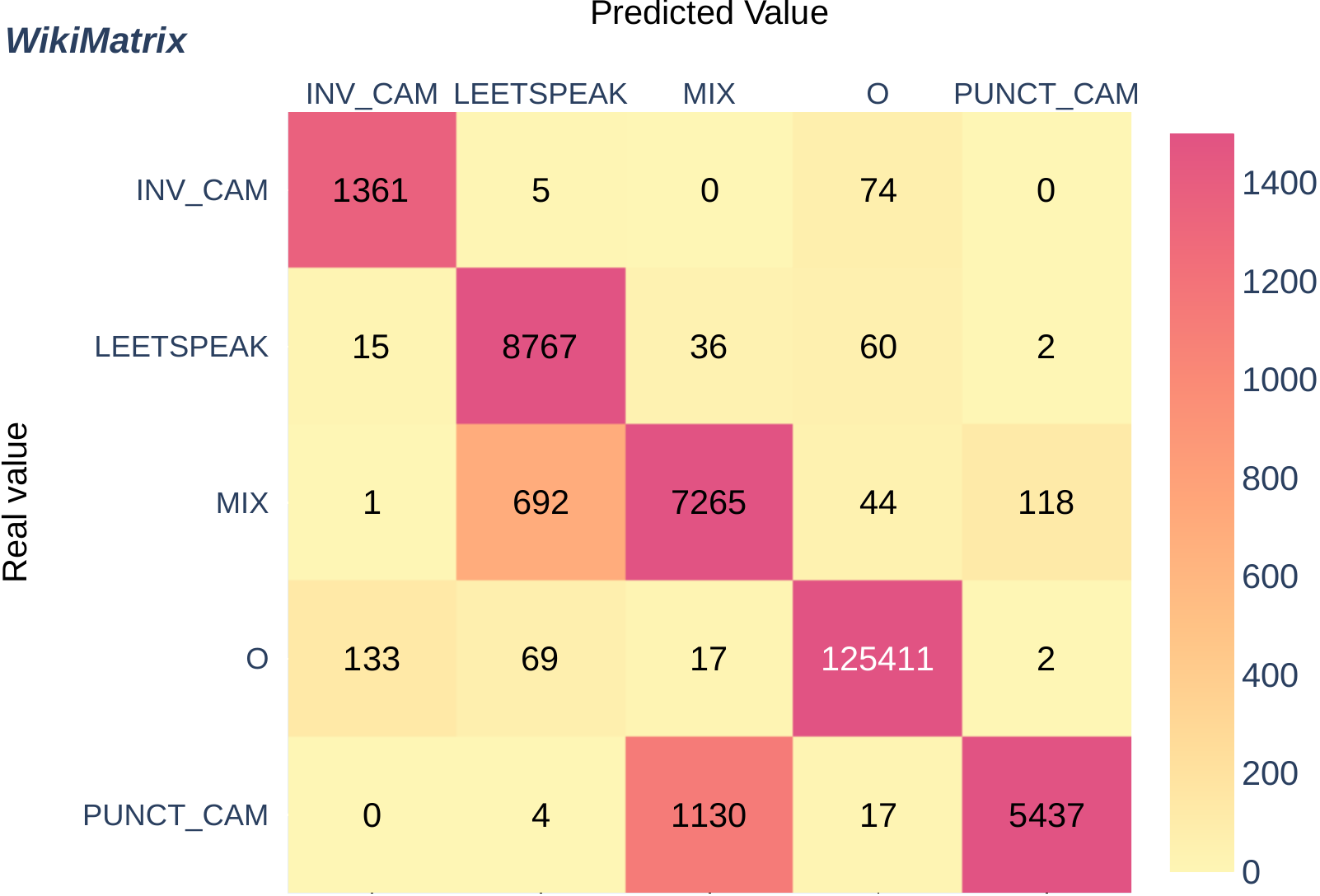}
\caption{}
\label{cm-WikiMatrix}
\end{subfigure}

\caption{Entity-level confusion matrix of word camouflage for the best multilingual NER model on (a) the whole multilingual test data. Results also broken down by dataset source: (b) News Commentary, (c) ParaCrawl, (d) TED2020, and (e) WikiMatrix. The rows represent the actual camouflage type, and in columns the type predicted by the model, where the number in each cell indicates the number of entities.}
\label{conf-matrix}
\end{figure}

The scores presented above for MPNET-ideal, the best multilingual model, are remarkable, as it achieves great results by detecting word camouflage and distinguishing the type of word camouflage technique used. To analyze in greater detail the capacity of the models to detect each of the different entities, the corresponding confusion matrix is also provided (Figure~\ref{conf-matrix}). It is important to note that the entity ``O' comes from ``Outside" and refers to those terms that are not entities to be detected.

As expected, one aspect of interest that emerged from the confusion matrices is that, across all datasets, trying to differentiate ``MIX" entities from ``LEETSPEAK" or ``PUNCT\_CAMO" entities is more complex as they are closely related. In the same way, these matrixes show that detecting inversion camouflage is harder than detecting punctuation or leetspeak camouflage as inversion shows more false positive and false negative with ``O" entity class.  

Taken together, these results suggest that developing Transformer models to detect word camouflage entities at a multilingual level is possible with great results in different tested scenarios.

\begin{figure}[H]
\centering
\begin{subfigure}[h]{0.48\linewidth}
\includegraphics[width=\linewidth]{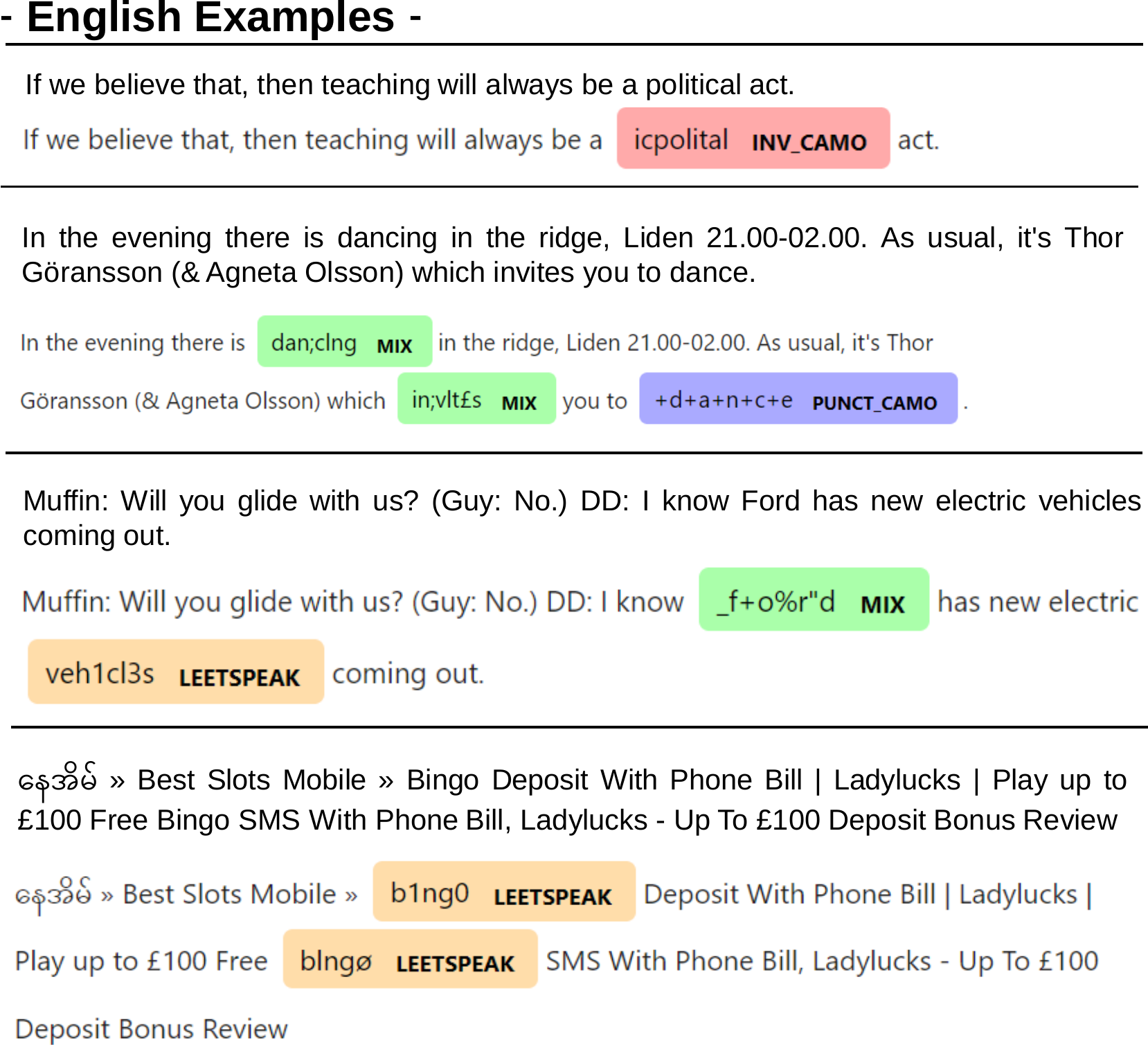}
\caption{}
\label{example-EN}
\end{subfigure}
\hspace{0.4cm}
\begin{subfigure}[h]{0.48\linewidth}
\includegraphics[width=\linewidth]{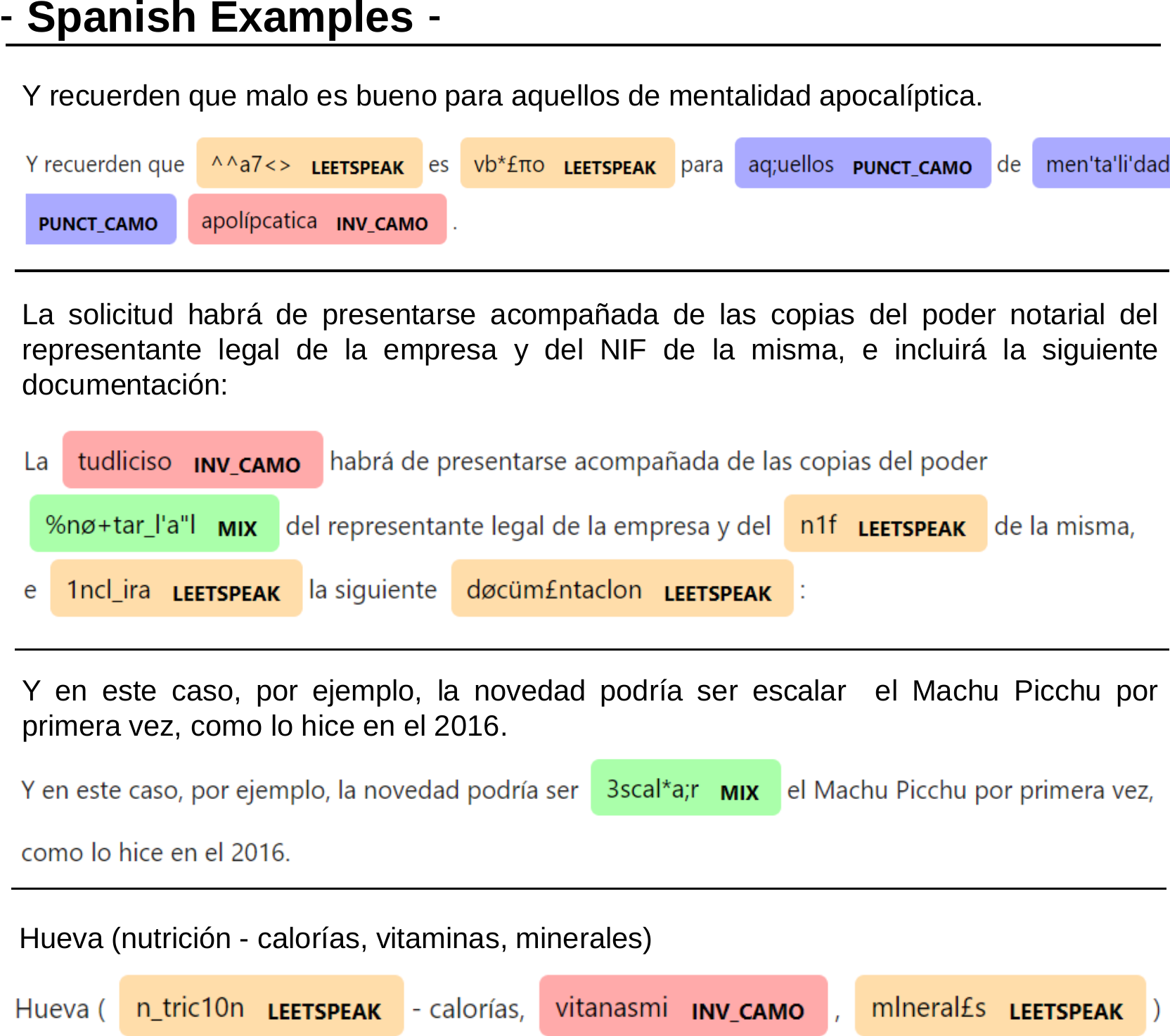}
\caption{}
\label{example-ES}
\end{subfigure}

\vspace{0.4cm}

\begin{subfigure}[h]{0.48\linewidth}
\includegraphics[width=\linewidth]{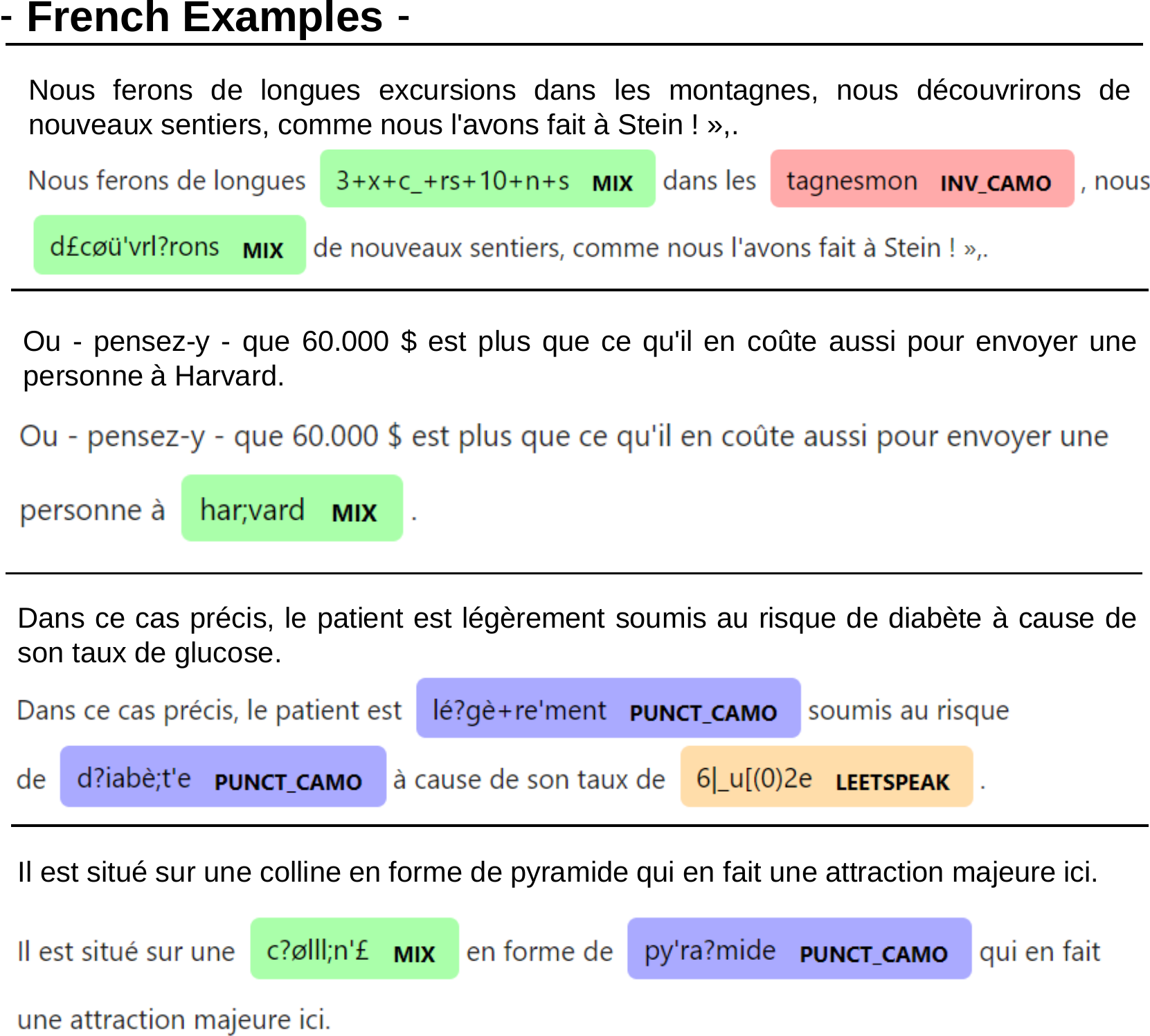}
\caption{}
\label{example-FR}
\end{subfigure}
\hspace{0.4cm}
\begin{subfigure}[h]{0.48\linewidth}
\includegraphics[width=\linewidth]{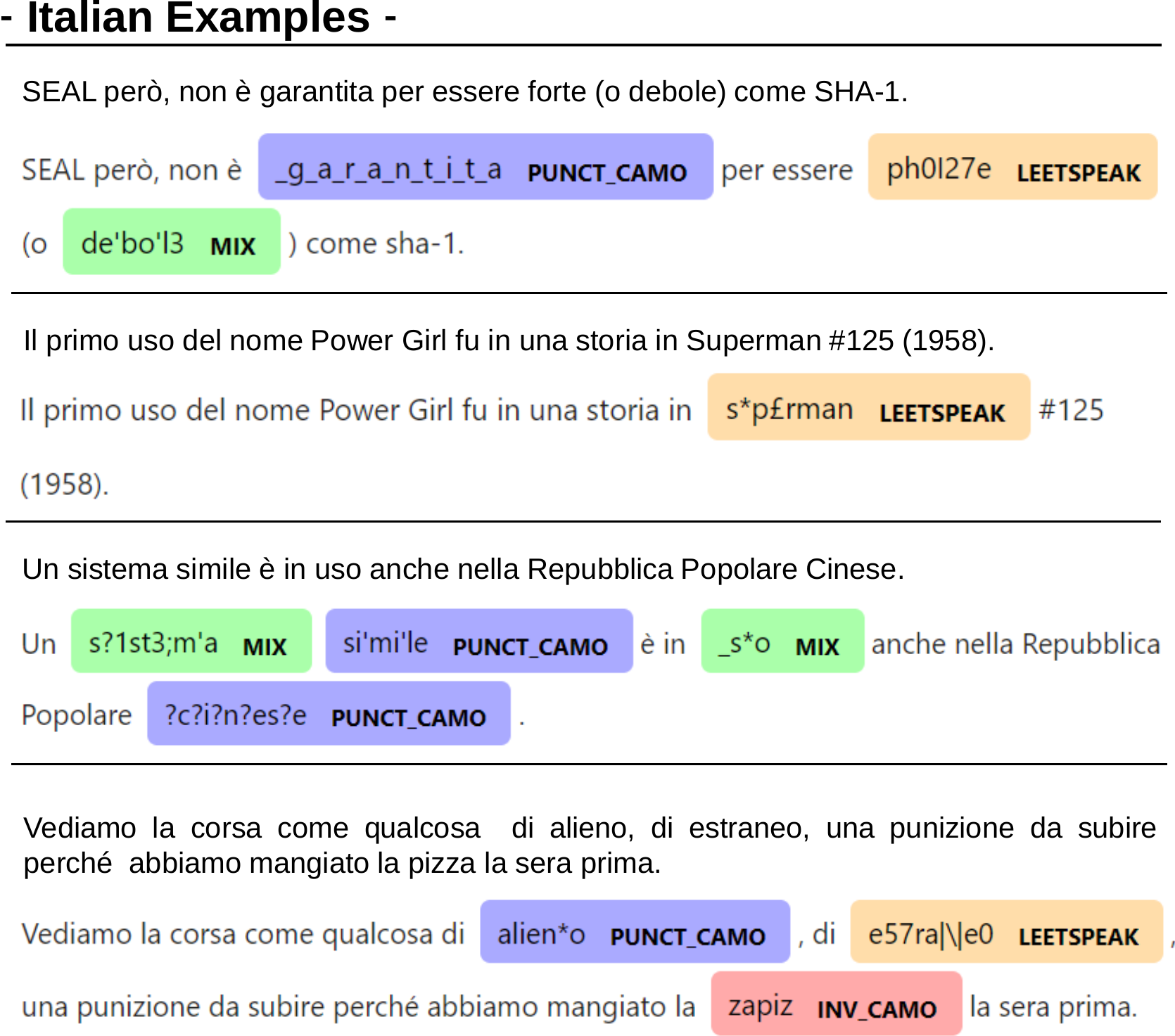}
\caption{}
\label{example-IT}
\end{subfigure}

\vspace{0.4cm}

\begin{subfigure}[h]{0.8\linewidth}
\includegraphics[width=\linewidth]{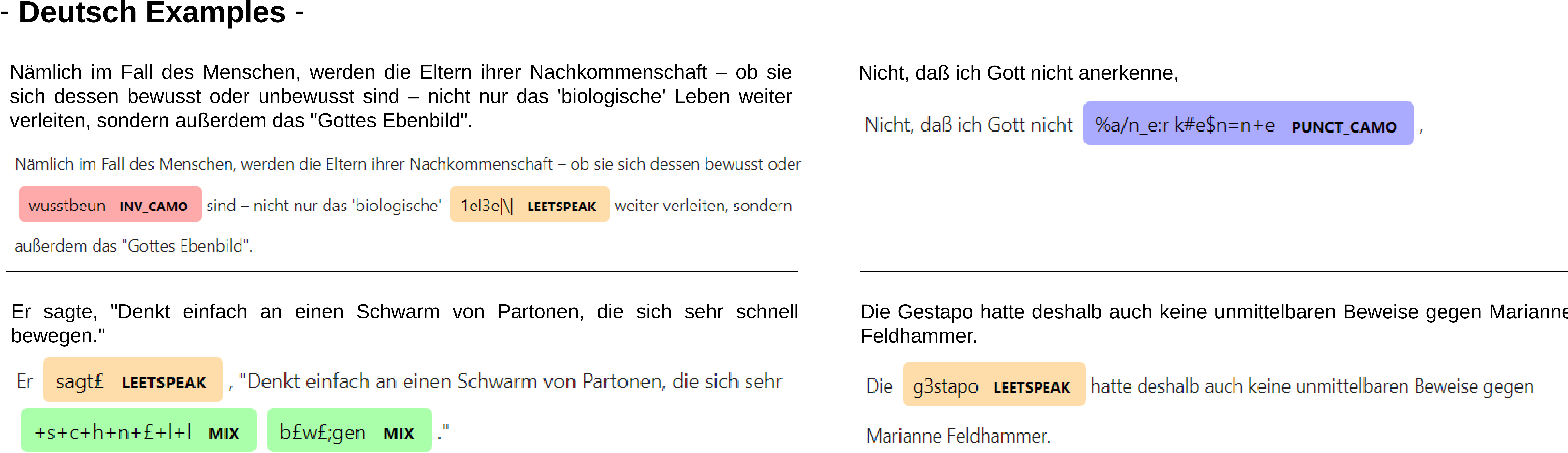}
\caption{}
\label{example-DE}
\end{subfigure}

\caption{Examples of word camouflage detection of the best multilingual NER model in (a) English, (b) Spanish, (c) French, (d) Italian and (e) German.}
\label{NER_examples}
\end{figure}

In addition, Figure \ref{NER_examples} shows some synthetic examples of the split test in the detection of camouflaged content. From this figure, we can see how the best multilingual model is able to detect and differentiate the different content avoidance strategies at multilingual levels from the simplest to the more complex word camouflaging strategies. Likewise, an application\footnote{\href{https://huggingface.co/spaces/Huertas97/LeetSpeak-NER}{\url{https://huggingface.co/spaces/Huertas97/LeetSpeak-NER}}} has been developed where the model can be tested.

Finally, real cases of the use of word camouflage have been incorporated to test the performance of the proposed multilingual model.

\section{Conclusions}
\label{sec:conclusions}


This work has focused on continuing previous research conducted at the 22nd International Conference on Intelligent Data Engineering and Automated Learning (IDEAL)~\cite{aida_model_2021} on the usefulness of developing datasets and multilingual solutions in the field of Natural Language Processing along with the recent emerging content evasion phenomenon. 

Content evasion involves modifying the wording or formatting of a text to avoid detection by automated systems or human moderators. Besides, evidence of its existence are easily to find on social networks\textsuperscript{\ref{foot:sel-harm},\ref{foot:incel},\ref{foot:violacion},\ref{foot:porno}}. 
Moreover, content evasion detection can be useful for organisations and individuals who want to monitor and filter out misinformation and terrorist, misogynistic, or hateful language content from their platforms or networks. The counter of content evasion is essential in the context of misinformation because it is a tactic that is often used by those who spread false or misleading information to avoid detection and removal of their content by social media platforms and other online services.  By detecting and removing content that has been evaded, it is possible to reduce the spread of misinformation and protect users from being exposed to harmful or misleading information. Furthermore, by cracking down on content evasion, it is possible to make it more difficult for those who spread misinformation or misbehave to continue doing so, which can help reduce the overall prevalence of misinformation online.

Considerable progress has been made in this work regarding simulating and detecting content evasion techniques. This work has presented the ``pyleetspeak'' Python package\textsuperscript{\ref{foot:pyleetspeak}}, a novel multilingual customisable tool to simulate content evasion techniques based on textual modification and camouflaging. Furthermore, this work also provides a curated synthetic multilingual dataset\textsuperscript{\ref{foot:link-data}} obtained by applying the word camouflaging simulator on texts from various resources. This publicly available tool and dataset provide an initial step to counter new information disorder from malicious actors in the co-evolving information warfare battleground from a multilingual point of view.

Finally, the above resources have been employed to develop a powerful multilingual NER camouflage detection model\textsuperscript{ \ref{foot:model}} that identifies different word camouflage techniques. Our experiments results have corroborated the previous research~\cite{aida_model_2021} by improving the perform of multilingual models in camouflaged detection when using a pre-train in multilingual Semantic Textual Similarity Benchmark (mSTSb). Additionally, the best multilingual model has been evaluated in five languages (English, Spanish, French, Italian and German) and has proved to outperforms with monolingual baseline models for the languages
considered.

This work has proved that it is 
possible to automatically address the challenging task of identifying and flagging content that contains word camouflaging. Considering potential applications and implications of this work and future work, we would like to suggest that the use of ``pyleetspeak'' is not limited to generating data for NER training word camouflage models; it can also be applied as a data augmentation tool to make current content-dependent AI systems more robust. Future studies will investigate the impact of word camouflage on the performance of models trained on canonical text datasets that are applied in real-world scenarios susceptible to word camouflage and content evasion. Furthermore, we propose future research examining how word camouflaging occurs in different circumstances. Data collection from other content evasion cases combined with word camouflage NER detection provides an excellent opportunity to gain insight into which words are most susceptible to being camouflaged and which customised camouflaging methods are applied in malicious communities.
Therefore, the tools presented in this paper might help to address the critical issue of author profiling.
Furthermore, although the proposed study has only been tested in English, Spanish, French and Italian and German, our results prove that our work is easily extensible to other languages and situations, as the tools developed currently support more than 20 languages and have obtained excellent results. Further implementations will cover different content evasion strategies (i.e., paralanguage, use of emoticons instead of original symbols).


\section*{Acknowledgements}
This research has been supported by the Spanish Ministry of Science and Education under FightDIS (PID2020-117263GB-I00) and XAI-Disinfodemics (PLEC2021-007681) grants, by Comunidad Aut\'{o}noma de Madrid under S2018/ TCS-4566 (CYNAMON), by BBVA Foundation grants for scientific research teams SARS-CoV-2 and COVID-19 under the grant: "\textit{CIVIC: Intelligent characterisation of the veracity of the information related to COVID-19}", and by IBERIFIER (Iberian Digital Media Research and Fact-Checking Hub), funded by the European Commission under the call CEF-TC-2020-2, grant number 2020-EU-IA-0252. Finally, David Camacho has been supported by the Comunidad Aut\'{o}noma de Madrid under "Convenio Plurianual with the Universidad Politécnica de Madrid in the actuation line of \textit{Programa de Excelencia para el Profesorado Universitario}"

\bibliographystyle{elsarticle-num}
\bibliography{references.bib}

\begin{thebibliography}{10}
\expandafter\ifx\csname url\endcsname\relax
  \def\url#1{\texttt{#1}}\fi
\expandafter\ifx\csname urlprefix\endcsname\relax\def\urlprefix{URL }\fi
\expandafter\ifx\csname href\endcsname\relax
  \def\href#1#2{#2} \def\path#1{#1}\fi

\bibitem{FaganFrank2020Osmc}
F.~Fagan, {Optimal social media content moderation and platform immunities},
  {European Journal of Law and Economics} {50}~({3, SI}) ({2020}) {437--449}.
\newblock \href {https://doi.org/10.1007/s10657-020-09653-7}
  {\path{doi:10.1007/s10657-020-09653-7}}.

\bibitem{content-filtering}
N.~Thilagavathi, R.~Taarika, Content based filtering in online social network
  using inference algorithm, in: 2014 International Conference on Circuits,
  Power and Computing Technologies [ICCPCT-2014], 2014, pp. 1416--1420.
\newblock \href {https://doi.org/10.1109/ICCPCT.2014.7054762}
  {\path{doi:10.1109/ICCPCT.2014.7054762}}.

\bibitem{beyond-content-2018}
Y.~Gerrard, {Beyond the hashtag: Circumventing content moderation on social
  media}, {New Media \& Society} {20}~({12}) ({2018}) {4492--4511}.
\newblock \href {https://doi.org/10.1177/1461444818776611}
  {\path{doi:10.1177/1461444818776611}}.

\bibitem{ads-in-osn}
L.~Kelly, G.~Kerr, J.~Drennan, Avoidance of advertising in social networking
  sites, Journal of Interactive Advertising 10~(2) (2010) 16--27.
\newblock \href {https://doi.org/10.1080/15252019.2010.10722167}
  {\path{doi:10.1080/15252019.2010.10722167}}.

\bibitem{instagram-post-study}
S.~Chancellor, J.~A. Pater, T.~Clear, E.~Gilbert, M.~De~Choudhury, \#thyghgapp:
  Instagram content moderation and lexical variation in pro-eating disorder
  communities, in: Proceedings of the 19th ACM Conference on Computer-Supported
  Cooperative Work \& Social Computing, CSCW '16, Association for Computing
  Machinery, New York, NY, USA, 2016, p. 1201–1213.
\newblock \href {https://doi.org/10.1145/2818048.2819963}
  {\path{doi:10.1145/2818048.2819963}}.

\bibitem{mosseri_addressing_2016}
A.~Mosseri,
  \href{https://about.fb.com/news/2016/12/news-feed-fyi-addressing-hoaxes-and-fake-news/}{Addressing
  {Hoaxes} and {Fake} {News}} (2016).
\newline\urlprefix\url{https://about.fb.com/news/2016/12/news-feed-fyi-addressing-hoaxes-and-fake-news/}

\bibitem{yoel__updating_nodate}
R.~Yoel, N.~Pickles,
  \href{https://blog.twitter.com/en_us/topics/product/2020/updating-our-approach-to-misleading-information}{Updating
  our approach to misleading information}.
\newline\urlprefix\url{https://blog.twitter.com/en_us/topics/product/2020/updating-our-approach-to-misleading-information}

\bibitem{SHAREVSKI2022102577}
F.~Sharevski, R.~Alsaadi, P.~Jachim, E.~Pieroni, Misinformation warnings:
  Twitter’s soft moderation effects on covid-19 vaccine belief echoes,
  Computers \& Security 114 (2022) 102577.
\newblock \href {https://doi.org/https://doi.org/10.1016/j.cose.2021.102577}
  {\path{doi:https://doi.org/10.1016/j.cose.2021.102577}}.

\bibitem{exsy-covid}
L.~S. Martinez, Health Misinformation and Rumors, John Wiley \& Sons, Ltd,
  2022, pp. 1--6.
\newblock \href
  {https://doi.org/https://doi.org/10.1002/9781119678816.iehc0950}
  {\path{doi:https://doi.org/10.1002/9781119678816.iehc0950}}.

\bibitem{noauthor_covid-19_nodate}
\href{https://developer.twitter.com/en/docs/labs/covid19-stream/overview}{{COVID}-19
  stream}, publisher: Twitter Developer Platform.
\newline\urlprefix\url{https://developer.twitter.com/en/docs/labs/covid19-stream/overview}

\bibitem{noauthor_twitter_nodate}
\href{https://developer.twitter.com/en/products/twitter-api/academic-research}{Twitter
  {API} for {Academic} {Research} {\textbar} {Products}}, publisher: Twitter
  Developer Platform.
\newline\urlprefix\url{https://developer.twitter.com/en/products/twitter-api/academic-research}

\bibitem{martin2021factercheck}
A.~Martín, J.~Huertas-Tato, {\'A}.~Huertas-García, G.~Villar-Rodríguez,
  D.~Camacho, {FacTeR-Check: Semi-automated fact-checking through semantic
  similarity and natural language inference}, Knowledge-Based Systems 251
  (2022) 109265.
\newblock \href {https://doi.org/10.1016/j.knosys.2022.109265}
  {\path{doi:10.1016/j.knosys.2022.109265}}.

\bibitem{noauthor_policy_YouTube}
\href{https://support.google.com/youtube/answer/9891785}{Policy on medical
  misinformation about {COVID}-19}, publisher: YouTube.
\newline\urlprefix\url{https://support.google.com/youtube/answer/9891785}

\bibitem{AI-content-2020}
R.~Gorwa, R.~Binns, C.~Katzenbach, Algorithmic content moderation: Technical
  and political challenges in the automation of platform governance (2020).
\newblock \href {https://doi.org/10.31235/osf.io/fj6pg}
  {\path{doi:10.31235/osf.io/fj6pg}}.

\bibitem{bridge_gap_leet_2005}
M.~Kavanagh, Bridge the generation gap by decoding leetspeak, Inside the
  Internet 12~(12) (2005) 11.

\bibitem{romero_wordcamo_2021}
A.~Romero-Vicente,
  \href{https://www.disinfo.eu/publications/word-camouflage-to-evade-content-moderation/}{Word
  camouflage to evade content moderation} (2021).
\newline\urlprefix\url{https://www.disinfo.eu/publications/word-camouflage-to-evade-content-moderation/}

\bibitem{aida_model_2021}
{\'A}.~Huertas-Garc{\'i}a, J.~Huertas-Tato, A.~Mart{\'i}n~Garc{\'i}a,
  D.~Camacho, Countering {Misinformation} {Through} {Semantic}-{Aware}
  {Multilingual} {Models}, in: Intelligent {Data} {Engineering} and {Automated}
  {Learning} – {IDEAL} 2021, Springer International Publishing, 2021, pp.
  312--323.
\newblock \href {https://doi.org/10.1007/978-3-030-91608-4_31}
  {\path{doi:10.1007/978-3-030-91608-4_31}}.

\bibitem{content_sex_moderation_2020}
Y.~Gerrard, H.~Thornham, {Content moderation: Social media's sexist
  assemblages}, {New Media \& Society} {22}~({7, SI}) ({2020}) {1266--1286}.
\newblock \href {https://doi.org/{10.1177/1461444820912540}}
  {\path{doi:{10.1177/1461444820912540}}}.

\bibitem{trad_moderation_004}
C.~Lampe, P.~Resnick, Slash(dot) and burn: Distributed moderation in a large
  online conversation space, in: Proceedings of the SIGCHI Conference on Human
  Factors in Computing Systems, CHI '04, Association for Computing Machinery,
  New York, NY, USA, 2004, p. 543–550.
\newblock \href {https://doi.org/10.1145/985692.985761}
  {\path{doi:10.1145/985692.985761}}.

\bibitem{Elkin-KorenNiva2020CaRt}
N.~Elkin-Koren, {Contesting algorithms: Restoring the public interest in
  content filtering by artificial intelligence}, {Big Data \& Society}
  {7}~({2}) ({2020}).
\newblock \href {https://doi.org/{10.1177/2053951720932296}}
  {\path{doi:{10.1177/2053951720932296}}}.

\bibitem{cobbe_algorithmic_2021}
J.~Cobbe, Algorithmic {Censorship} by {Social} {Platforms}: {Power} and
  {Resistance}, Philosophy \& Technology 34~(4) (2021) 739--766.
\newblock \href {https://doi.org/10.1007/s13347-020-00429-0}
  {\path{doi:10.1007/s13347-020-00429-0}}.

\bibitem{sumpter_outnumbered_2018}
D.~Sumpter, Outnumbered: from {Facebook} and {Google} to fake news and
  filter-bubbles - the algorithms that control our lives, Bloomsbury Sigma,
  London, 2018, oCLC: on1035374425.

\bibitem{ofcom_use_2019}
{Ofcom},
  \href{https://www.cambridgeconsultants.com/insights/whitepaper/ofcom-use-ai-online-content-moderation}{Use
  of {AI} in online content moderation}, publisher: Cambridge Consultants
  (2019).
\newline\urlprefix\url{https://www.cambridgeconsultants.com/insights/whitepaper/ofcom-use-ai-online-content-moderation}

\bibitem{youtube_ban_results}
\href{https://perma.cc/44V5-554U}{Global {Internet} {Forum} to {Counter}
  {Terrorism} {\textbar} {About}}.
\newline\urlprefix\url{https://perma.cc/44V5-554U}

\bibitem{ferreira_antivaccine_2020}
F.~Ferreira, Antivaccine videos slip through {YouTube}’s advertising
  policies, new study finds, Science (2020).
\newblock \href {https://doi.org/10.1126/science.abf5402}
  {\path{doi:10.1126/science.abf5402}}.

\bibitem{_twitter_inc_permanent_2021}
{ Twitter Inc},
  \href{https://blog.twitter.com/en_us/topics/company/2020/suspension}{Permanent
  suspension of @{realDonaldTrump}} (2021).
\newline\urlprefix\url{https://blog.twitter.com/en_us/topics/company/2020/suspension}

\bibitem{twitter_blog_nuevo_2022}
T.~Blog,
  \href{https://blog.twitter.com/es_es/topics/2022/nuevo-canal-para-reportar-informacion-potencialmente-enganosa-en}{Nuevo
  canal para reportar información potencialmente engañosa en {Twitter}}
  (2022).
\newline\urlprefix\url{https://blog.twitter.com/es_es/topics/2022/nuevo-canal-para-reportar-informacion-potencialmente-enganosa-en}

\bibitem{_bickert_removing_2020}
M.~Bickert,
  \href{https://about.fb.com/news/2020/10/removing-holocaust-denial-content/}{Removing
  {Holocaust} {Denial} {Content}} (2020).
\newline\urlprefix\url{https://about.fb.com/news/2020/10/removing-holocaust-denial-content/}

\bibitem{phising_detection_ml}
U.~Ozker, O.~K. Sahingoz, Content based phishing detection with machine
  learning, in: 2020 International Conference on Electrical Engineering (ICEE),
  2020, pp. 1--6.
\newblock \href {https://doi.org/10.1109/ICEE49691.2020.9249892}
  {\path{doi:10.1109/ICEE49691.2020.9249892}}.

\bibitem{wall_filter}
N.~Thilagavathi, R.~Taarika, Content based filtering in online social network
  using inference algorithm, in: 2014 International Conference on Circuits,
  Power and Computing Technologies [ICCPCT-2014], 2014, pp. 1416--1420.
\newblock \href {https://doi.org/10.1109/ICCPCT.2014.7054762}
  {\path{doi:10.1109/ICCPCT.2014.7054762}}.

\bibitem{user_wall_2}
A.~S. Vairagade, R.~A. Fadnavis, Automated content based short text
  classification for filtering undesired posts on facebook, in: 2016 World
  Conference on Futuristic Trends in Research and Innovation for Social Welfare
  (Startup Conclave), 2016, pp. 1--5.
\newblock \href {https://doi.org/10.1109/STARTUP.2016.7583984}
  {\path{doi:10.1109/STARTUP.2016.7583984}}.

\bibitem{Blashki2005GameGG}
K.~Blashki, S.~Nichol, Game geek's goss: linguistic creativity in young males
  within an online university forum (94$/\backslash/\backslash$3 933k’5
  9055oneone), 2005.

\bibitem{Shaari2015NetspeakAA}
A.~H. Shaari, K.~B.~A. Bataineh, Netspeak and a breach of formality:
  Informalization and fossilization of errors in writing among esl and efl
  learners, International Journal for Cross-Disciplinary Subjects in Education
  6 (2015) 2165--2173.

\bibitem{passwords_leet_2019}
J.~Kavrestad, F.~Eriksson, M.~Nohlberg, {Understanding passwords - a taxonomy
  of password creation strategies}, {Information and Computer Security}
  {27}~({3}) ({2019}) {453--467}.
\newblock \href {https://doi.org/{10.1108/ICS-06-2018-0077}}
  {\path{doi:{10.1108/ICS-06-2018-0077}}}.

\bibitem{fuchs__2013}
J.~Fuchs,
  \href{https://unipub.uni-graz.at/obvugrhs/content/titleinfo/231890?lang=en}{{Gamespeak}
  for n00bs - a linguistic and pragmatic analysis of gamers' language}, Ph.D.
  thesis, University of Graz (2013).
\newline\urlprefix\url{https://unipub.uni-graz.at/obvugrhs/content/titleinfo/231890?lang=en}

\bibitem{password_leet_use_2016}
M.~Golla, B.~Beuscher, M.~Duermuth, {On the Security of Cracking-Resistant
  Password Vaults}, in: {CCS'16: Proceddings of the 2016 ACM SIGSAC Conference
  on Computer and Comunication Security}, {2016}, pp. {1230--1241}.
\newblock \href {https://doi.org/{10.1145/2976749.2978416}}
  {\path{doi:{10.1145/2976749.2978416}}}.

\bibitem{password_dropbox_2016}
D.~L. Wheeler, Zxcvbn: Low-budget password strength estimation, in: Proceedings
  of the 25th USENIX Conference on Security Symposium, SEC'16, USENIX
  Association, USA, 2016, p. 157–173.

\bibitem{password_leet_2021}
K.~H. Hong, U.~G. Kang, B.~M. Lee, {Enhanced Evaluation Model of Security
  Strength for Passwords Using Integrated Korean and English Password
  Dictionaries}, {Security and Communication Networks} {2021} ({2021}).
\newblock \href {https://doi.org/{10.1155/2021/3122627}}
  {\path{doi:{10.1155/2021/3122627}}}.

\bibitem{noauthor_cybersquatting_2019}
\href{https://www.incibe.es/protege-tu-empresa/blog/cybersquatting-y-protegerse}{Cybersquatting,
  qué es y cómo protegerse} (2019).
\newline\urlprefix\url{https://www.incibe.es/protege-tu-empresa/blog/cybersquatting-y-protegerse}

\bibitem{spam_leet_2018}
W.~Peng, L.~Huang, J.~Jia, E.~Ingram, Enhancing the naive bayes spam filter
  through intelligent text modification detection, in: 2018 17th IEEE
  International Conference On Trust, Security And Privacy In Computing And
  Communications/ 12th IEEE International Conference On Big Data Science And
  Engineering (TrustCom/BigDataSE), 2018, pp. 849--854.
\newblock \href {https://doi.org/10.1109/TrustCom/BigDataSE.2018.00122}
  {\path{doi:10.1109/TrustCom/BigDataSE.2018.00122}}.

\bibitem{slang_media_2016}
T.~Singh, M.~Kumari, Role of text pre-processing in twitter sentiment analysis,
  Procedia Computer Science 89 (2016) 549--554.
\newblock \href {https://doi.org/https://doi.org/10.1016/j.procs.2016.06.095}
  {\path{doi:https://doi.org/10.1016/j.procs.2016.06.095}}.

\bibitem{slang_media_2020}
Z.~Z. Izazi, T.~M. Tengku-Sepora, {Slangs on Social Media: Variations among
  Malay Language Users on Twitter}, {Pertanika Journal of Social Science and
  Humanities} {28}~({1}) ({2020}) {17--34}.

\bibitem{Moskalenko_González_Kates_Morton_2022}
S.~Moskalenko, J.~F.-G. González, N.~Kates, J.~Morton, Incel ideology,
  radicalization and mental health: A survey study, The Journal of
  Intelligence, Conflict, and Warfare 4~(3) (2022) 1–29.
\newblock \href {https://doi.org/10.21810/jicw.v4i3.3817}
  {\path{doi:10.21810/jicw.v4i3.3817}}.

\bibitem{craenen_leet_cheatsheet}
R.~Craenen, \href{https://www.gamehouse.com/blog/leet-speak-cheat-sheet/}{Leet
  speak cheat sheet}.
\newline\urlprefix\url{https://www.gamehouse.com/blog/leet-speak-cheat-sheet/}

\bibitem{plotly}
P.~T. Inc., \href{https://plot.ly}{Collaborative data science} (2015).
\newline\urlprefix\url{https://plot.ly}

\bibitem{grootendorst2020keybert}
M.~Grootendorst, Keybert: Minimal keyword extraction with bert. (2020).
\newblock \href {https://doi.org/10.5281/zenodo.4461265}
  {\path{doi:10.5281/zenodo.4461265}}.

\bibitem{montani_spacy_2020}
I.~Montani, M.~Honnibal, S.~Van~Landeghem, A.~Boyd, "{spaCy}:
  {Industrial}-strength {Natural} {Language} {Processing} in {Python}" (2020).
\newblock \href {https://doi.org/10.5281/zenodo.1212303}
  {\path{doi:10.5281/zenodo.1212303}}.

\bibitem{vaswani2017attention}
A.~Vaswani, N.~Shazeer, N.~Parmar, J.~Uszkoreit, L.~Jones, A.~N. Gomez,
  L.~Kaiser, I.~Polosukhin, Attention is all you need (2017).
\newblock \href {http://arxiv.org/abs/1706.03762} {\path{arXiv:1706.03762}}.

\bibitem{devlin2019bert}
J.~Devlin, M.-W. Chang, K.~Lee, K.~Toutanova, Bert: Pre-training of deep
  bidirectional transformers for language understanding (2019).
\newblock \href {http://arxiv.org/abs/1810.04805} {\path{arXiv:1810.04805}}.

\bibitem{wolf-etal-2020-transformers}
T.~Wolf, L.~Debut, V.~Sanh, J.~Chaumond, C.~Delangue, A.~Moi, P.~Cistac,
  T.~Rault, R.~Louf, M.~Funtowicz, J.~Davison, S.~Shleifer, P.~von Platen,
  C.~Ma, Y.~Jernite, J.~Plu, C.~Xu, T.~L. Scao, S.~Gugger, M.~Drame, Q.~Lhoest,
  A.~M. Rush, Transformers: State-of-the-art natural language processing, in:
  Proceedings of the 2020 Conference on Empirical Methods in Natural Language
  Processing: System Demonstrations, Association for Computational Linguistics,
  2020, pp. 38--45.

\bibitem{opus_news_data}
J.~Tiedemann, Parallel data, tools and interfaces in opus, in: N.~C.~C. Chair),
  K.~Choukri, T.~Declerck, M.~U. Dogan, B.~Maegaard, J.~Mariani, J.~Odijk,
  S.~Piperidis (Eds.), Proceedings of the Eight International Conference on
  Language Resources and Evaluation (LREC'12), European Language Resources
  Association (ELRA), Istanbul, Turkey, 2012.

\bibitem{tiedemann-2012-parallel}
J.~Tiedemann, Parallel data, tools and interfaces in {OPUS}, in: Proceedings of
  the Eighth International Conference on Language Resources and Evaluation
  ({LREC}'12), European Language Resources Association (ELRA), Istanbul,
  Turkey, 2012, pp. 2214--2218.

\bibitem{banon-etal-2020-paracrawl}
M.~Ba{\~n}{\'o}n, P.~Chen, B.~Haddow, K.~Heafield, H.~Hoang,
  M.~Espl{\`a}-Gomis, M.~L. Forcada, A.~Kamran, F.~Kirefu, P.~Koehn,
  S.~Ortiz~Rojas, L.~Pla~Sempere, G.~Ram{\'\i}rez-S{\'a}nchez, E.~Sarr{\'\i}as,
  M.~Strelec, B.~Thompson, W.~Waites, D.~Wiggins, J.~Zaragoza, {P}ara{C}rawl:
  Web-scale acquisition of parallel corpora, in: Proceedings of the 58th Annual
  Meeting of the Association for Computational Linguistics, Association for
  Computational Linguistics, Online, 2020, pp. 4555--4567.
\newblock \href {https://doi.org/10.18653/v1/2020.acl-main.417}
  {\path{doi:10.18653/v1/2020.acl-main.417}}.

\bibitem{reimers_2020_multilingual_sentence_bert}
N.~Reimers, I.~Gurevych, Making monolingual sentence embeddings multilingual
  using knowledge distillation, arXiv preprint arXiv:2004.09813 (2020).

\bibitem{schwenk2019wikimatrix}
H.~Schwenk, V.~Chaudhary, S.~Sun, H.~Gong, F.~Guzmán, Wikimatrix: Mining 135m
  parallel sentences in 1620 language pairs from wikipedia (2019).
\newblock \href {http://arxiv.org/abs/1907.05791} {\path{arXiv:1907.05791}}.

\bibitem{song2020mpnet}
K.~Song, X.~Tan, T.~Qin, J.~Lu, T.-Y. Liu, Mpnet: Masked and permuted
  pre-training for language understanding (2020).
\newblock \href {http://arxiv.org/abs/2004.09297} {\path{arXiv:2004.09297}}.

\bibitem{cer-etal-2017-semeval}
D.~Cer, M.~Diab, E.~Agirre, I.~Lopez-Gazpio, L.~Specia, {S}em{E}val-2017 task
  1: Semantic textual similarity multilingual and crosslingual focused
  evaluation, in: Proceedings of the 11th International Workshop on Semantic
  Evaluation ({S}em{E}val-2017), Association for Computational Linguistics,
  Vancouver, Canada, 2017, pp. 1--14.

\bibitem{bloomz}
N.~Muennighoff, T.~Wang, L.~Sutawika, A.~Roberts, S.~Biderman, T.~L. Scao,
  M.~S. Bari, S.~Shen, Z.-X. Yong, H.~Schoelkopf, X.~Tang, D.~Radev, A.~F. Aji,
  K.~Almubarak, S.~Albanie, Z.~Alyafeai, A.~Webson, E.~Raff, C.~Raffel,
  Crosslingual generalization through multitask finetuning (2022).
\newblock \href {https://doi.org/10.48550/ARXIV.2211.01786}
  {\path{doi:10.48550/ARXIV.2211.01786}}.

\bibitem{scao2022bloom}
T.~L. Scao, A.~Fan, C.~Akiki, E.~Pavlick, S.~Ili{\'c}, D.~Hesslow,
  R.~Castagn{\'e}, A.~S. Luccioni, F.~Yvon, M.~Gall{\'e}, et~al., Bloom: A
  176b-parameter open-access multilingual language model, arXiv preprint
  arXiv:2211.05100 (2022).

\bibitem{mt5}
L.~Xue, N.~Constant, A.~Roberts, M.~Kale, R.~Al-Rfou, A.~Siddhant, A.~Barua,
  C.~Raffel, mt5: A massively multilingual pre-trained text-to-text transformer
  (2020).
\newblock \href {https://doi.org/10.48550/ARXIV.2010.11934}
  {\path{doi:10.48550/ARXIV.2010.11934}}.

\bibitem{conneau2020unsupervised}
A.~Conneau, K.~Khandelwal, N.~Goyal, V.~Chaudhary, G.~Wenzek, F.~Guzmán,
  E.~Grave, M.~Ott, L.~Zettlemoyer, V.~Stoyanov, {Unsupervised Cross-lingual
  Representation Learning at Scale} (2019).
\newblock \href {https://doi.org/10.48550/ARXIV.1911.02116}
  {\path{doi:10.48550/ARXIV.1911.02116}}.

\bibitem{liu2019roberta}
Z.~Liu, W.~Lin, Y.~Shi, J.~Zhao, {A Robustly Optimized BERT Pre-Training
  Approach with Post-Training}, in: Chinese Computational Linguistics: 20th
  China National Conference, CCL 2021, Hohhot, China, August 13–15, 2021,
  Proceedings, Springer-Verlag, Berlin, Heidelberg, 2021, p. 471–484.
\newblock \href {https://doi.org/10.1007/978-3-030-84186-7_31}
  {\path{doi:10.1007/978-3-030-84186-7_31}}.

\bibitem{devlin2018bert}
J.~Devlin, M.-W. Chang, K.~Lee, K.~Toutanova, {BERT: Pre-training of Deep
  Bidirectional Transformers for Language Understanding}, in: Proceedings of
  the 2019 Conference of the North {A}merican Chapter of the Association for
  Computational Linguistics: Human Language Technologies, Volume 1, Association
  for Computational Linguistics, Minneapolis, Minnesota, 2019, pp. 4171--4186.
\newblock \href {https://doi.org/10.18653/v1/N19-1423}
  {\path{doi:10.18653/v1/N19-1423}}.

\bibitem{Zhu_2015_ICCV}
Y.~Zhu, R.~Kiros, R.~Zemel, R.~Salakhutdinov, R.~Urtasun, A.~Torralba,
  S.~Fidler, Aligning books and movies: Towards story-like visual explanations
  by watching movies and reading books, in: The IEEE International Conference
  on Computer Vision (ICCV), 2015.

\bibitem{gutierrezfandino2021spanish}
A.~Gutiérrez-Fandiño, J.~Armengol-Estapé, M.~Pàmies, J.~Llop-Palao,
  J.~Silveira-Ocampo, C.~P. Carrino, A.~Gonzalez-Agirre, C.~Armentano-Oller,
  C.~Rodriguez-Penagos, M.~Villegas, Spanish language models (2021).
\newblock \href {http://arxiv.org/abs/2107.07253} {\path{arXiv:2107.07253}}.

\bibitem{martin2020camembert}
L.~Martin, B.~Muller, P.~J.~O. Su{\'a}rez, Y.~Dupont, L.~Romary, {\'E}.~V.
  de~la Clergerie, D.~Seddah, B.~Sagot, Camembert: a tasty french language
  model, in: Proceedings of the 58th Annual Meeting of the Association for
  Computational Linguistics, 2020.

\bibitem{OSCAR}
P.~J.~O. Suárez, B.~Sagot, L.~Romary, Asynchronous pipelines for processing
  huge corpora on medium to low resource infrastructures, Proceedings of the
  Workshop on Challenges in the Management of Large Corpora (CMLC-7) 2019.
  Cardiff, 22nd July 2019, Leibniz-Institut f{\"u}r Deutsche Sprache, Mannheim,
  2019, pp. 9 -- 16.
\newblock \href {https://doi.org/10.14618/ids-pub-9021}
  {\path{doi:10.14618/ids-pub-9021}}.

\bibitem{gottbert}
R.~Scheible, F.~Thomczyk, P.~Tippmann, V.~Jaravine, M.~Boeker, Gottbert: a pure
  german language model (2020).
\newblock \href {https://doi.org/10.48550/ARXIV.2012.02110}
  {\path{doi:10.48550/ARXIV.2012.02110}}.

\end{thebibliography}

\newpage
\appendix

\section{Real Examples}
\label{}

\begin{figure}[!htpb]
\captionsetup[subfigure]{labelformat=empty}
\centering
\begin{subfigure}[h]{0.46\linewidth}
\includegraphics[width=\linewidth]{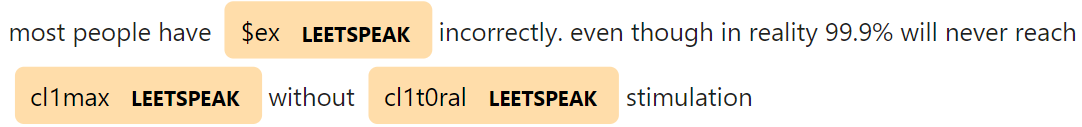}
\caption{\href{https://www.tiktok.com/@yuvalmann.s/video/7161061303723904261}{\url{Link} to the resource}}
\label{real-example-1}
\end{subfigure}
\hspace{0.2cm}
\begin{subfigure}[h]{0.46\linewidth}
\includegraphics[width=\linewidth]{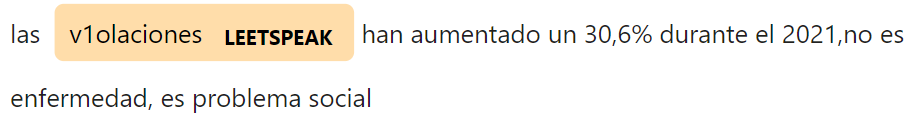}
\caption{\href{https://www.tiktok.com/@ladyxynthia/video/7058302808633675014}{\url{Link} to the resource}}
\label{real-example-2}
\end{subfigure}

\vspace{1cm}

\centering
\begin{subfigure}[h]{0.46\linewidth}
\includegraphics[width=\linewidth]{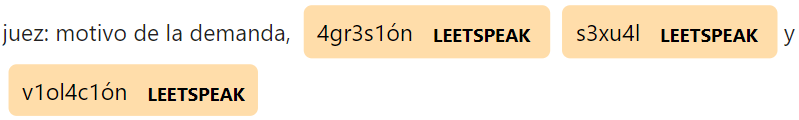}
\caption{\href{https://www.tiktok.com/@eri_gmz_03/video/6989646818015546630}{\url{Link} to the resource}}
\label{real-example-3}
\end{subfigure}
\hspace{0.2cm}
\begin{subfigure}[h]{0.46\linewidth}
\includegraphics[width=\linewidth]{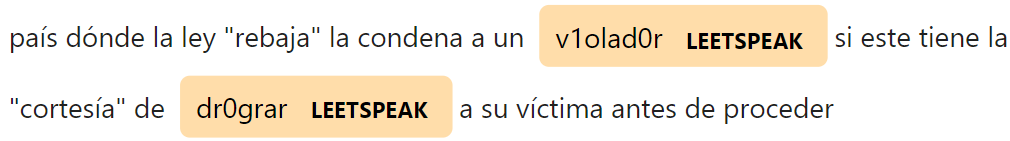}
\caption{\href{https://www.tiktok.com/@lydarte/video/7125192734784228613}{\url{Link} to the resource}}
\label{real-example-5}
\end{subfigure}

\vspace{1cm}

\begin{subfigure}[h]{0.46\linewidth}
\includegraphics[width=\linewidth]{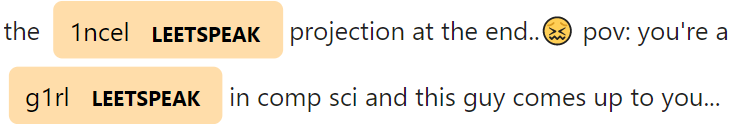}
\caption{\href{https://www.tiktok.com/@thebenjohns/video/7057624359942262021}{\url{Link} to the resource}}
\label{real-example-6}
\end{subfigure}
\hspace{0.4cm}
\begin{subfigure}[h]{0.46\linewidth}
\includegraphics[width=\linewidth]{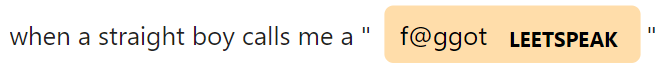}
\caption{\href{https://www.tiktok.com/@alfienichollsx/video/7014172606882532614}{\url{Link} to the resource}}
\label{real-example-6}
\end{subfigure}

\vspace{1cm}

\begin{subfigure}[h]{0.46\linewidth}
\includegraphics[width=\linewidth]{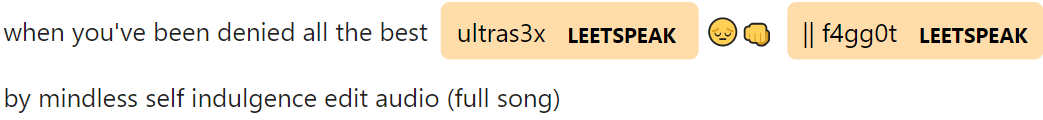}
\caption{\href{https://www.tiktok.com/@_k1tt13/video/7155896474918784262}{\url{Link} to the resource}}
\label{real-example-8}
\end{subfigure}
\hspace{0.2cm}
\begin{subfigure}[h]{0.46\linewidth}
\includegraphics[width=\linewidth]{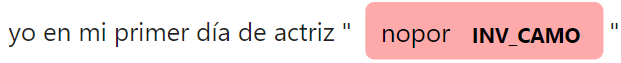}
\caption{\href{https://www.tiktok.com/@nataliaquirozl/video/6949722515195317510}{\url{Link} to the resource}}
\label{real-example-9}
\end{subfigure}

\vspace{1cm}

\begin{subfigure}[h]{0.3\linewidth}
\includegraphics[width=\linewidth]{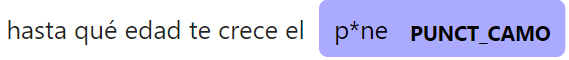}
\caption{\href{https://www.tiktok.com/@dr.pabloofficial/video/7016481297363242245}{\url{Link} to the resource}}
\label{real-example-10}
\end{subfigure}
\hspace{4cm}
\begin{subfigure}[h]{0.3\linewidth}
\includegraphics[width=\linewidth]{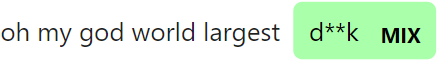}
\caption{\href{https://www.tiktok.com/@yari.ceny/video/6971084394991815942}{\url{Link} to the resource}}
\label{real-example-11}
\end{subfigure}

\vspace{0.4cm}
\begin{subfigure}[h]{0.3\linewidth}
\includegraphics[width=\linewidth]{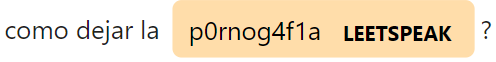}
\caption{\href{https://www.tiktok.com/@isaacquirogas/video/7152608826213141765}{\url{Link} to the resource}}
\label{real-example-12}
\end{subfigure}
\hspace{4cm}
\centering
\begin{subfigure}[h]{0.3\linewidth}
\includegraphics[width=\linewidth]{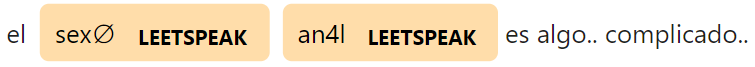}
\caption{\href{https://www.tiktok.com/@kstzuzu/video/7162939686602755334}{\url{Link} to the resource}}
\label{real-example-13}
\end{subfigure}

\vspace{1cm}

\begin{subfigure}[h]{0.46\linewidth}
\includegraphics[width=\linewidth]{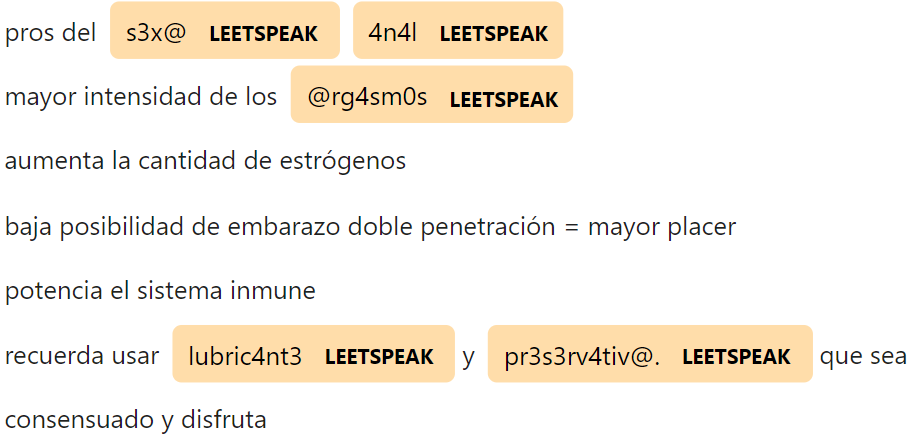}
\caption{\href{https://www.tiktok.com/@dra.carolinam/video/7027559383206792454}{\url{Link} to the resource}}
\label{real-example-14}
\end{subfigure}
\hspace{1cm}
\begin{subfigure}[h]{0.3\linewidth}
\includegraphics[width=\linewidth]{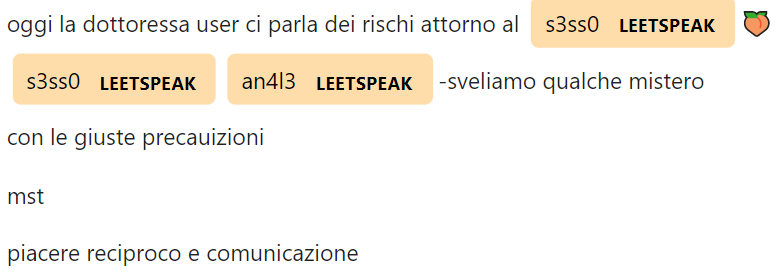}
\caption{\href{https://www.tiktok.com/@ciao_joyclub/video/7158124597869595909}{\url{Link} to the resource}}
\label{real-example-15}
\end{subfigure}
\end{figure}

\begin{figure}[t!]
\captionsetup[subfigure]{labelformat=empty}
\centering
\begin{subfigure}[h]{0.46\linewidth}
\includegraphics[width=\linewidth]{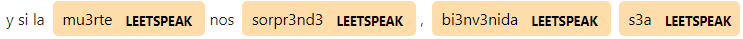}
\caption{\href{https://twitter.com/star_momonga/status/1600923223645790208}{\url{Link} to the resource}}
\label{real-example-16}
\end{subfigure}
\hspace{1cm}
\begin{subfigure}[h]{0.46\linewidth}
\includegraphics[width=\linewidth]{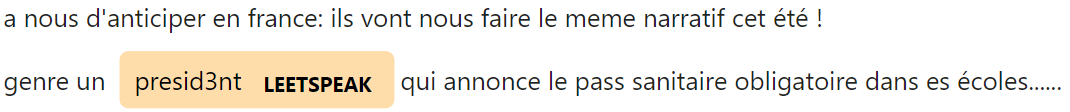}
\caption{\href{}{\url{Link} to the resource}}
\label{real-example-17}
\end{subfigure}

\vspace{1cm}

\centering
\begin{subfigure}[t]{0.46\linewidth}
\includegraphics[width=\linewidth]{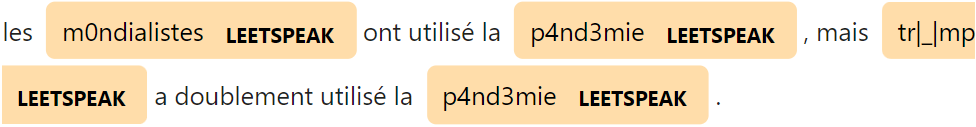}
\caption{\href{https://twitter.com/CarolineLessar8/status/1482908038252089344}{\url{Link} to the resource}}
\label{real-example-18}
\end{subfigure}
\hspace{1cm}
\begin{subfigure}[t]{0.46\linewidth}
\includegraphics[width=\linewidth]{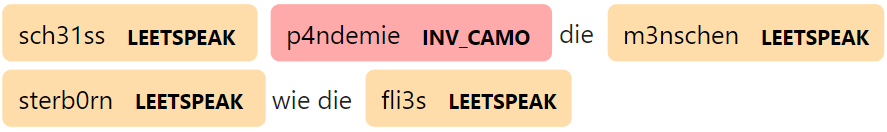}
\caption{\href{https://twitter.com/dan1eLLL/status/1397439491661606912}{\url{Link} to the resource}}
\label{real-example-19}
\end{subfigure}

\vspace{1cm}

\centering
\begin{subfigure}[t]{0.46\linewidth}
\includegraphics[width=\linewidth]{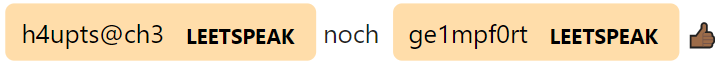}
\caption{\href{https://twitter.com/dan1eLLL/status/1397448393371312128}{\url{Link} to the resource}}
\label{real-example-20}
\end{subfigure}
\hspace{1cm}
\begin{subfigure}[t]{0.46\linewidth}
\includegraphics[width=\linewidth]{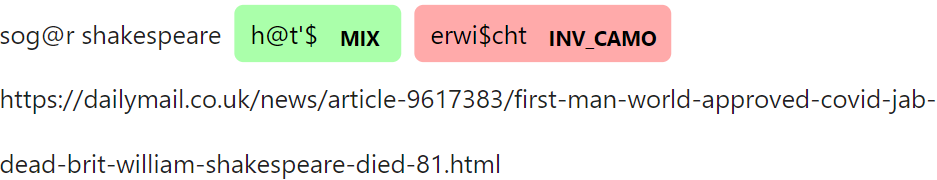}
\caption{\href{https://twitter.com/djmazzafakka/status/1397447497312780289}{\url{Link} to the resource}}
\label{real-example-21}
\end{subfigure}

\vspace{1cm}

\centering
\begin{subfigure}[t]{0.46\linewidth}
\includegraphics[width=\linewidth]{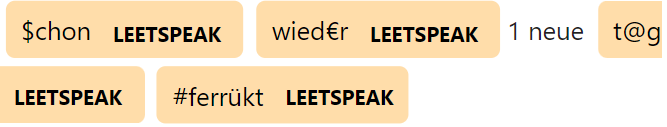}
\caption{\href{https://twitter.com/djmazzafakka/status/1397433511158632449}{\url{Link} to the resource}}
\label{real-example-22}
\end{subfigure}
\hspace{1cm}
\begin{subfigure}[t]{0.46\linewidth}
\includegraphics[width=\linewidth]{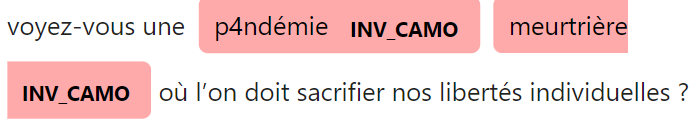}
\caption{\href{https://twitter.com/jauwanne/status/1493758654977167361}{\url{Link} to the resource}}
\label{real-example-23}
\end{subfigure}

\vspace{1cm}

\centering
\begin{subfigure}[t]{0.46\linewidth}
\includegraphics[width=\linewidth]{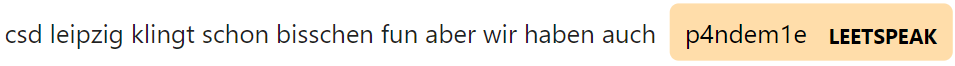}
\caption{\href{https://twitter.com/igelmithut/status/1414594729132630020}{\url{Link} to the resource}}
\label{real-example-24}
\end{subfigure}
\hspace{1cm}
\begin{subfigure}[t]{0.46\linewidth}
\includegraphics[width=\linewidth]{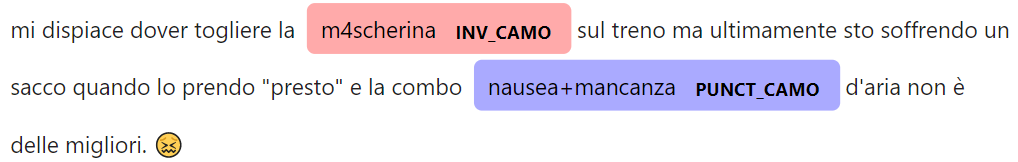}
\caption{\href{https://twitter.com/st1la_/status/1588425645071089664}{\url{Link} to the resource}}
\label{real-example-25}
\end{subfigure}

\vspace{1cm}

\centering
\begin{subfigure}[t]{0.46\linewidth}
\includegraphics[width=\linewidth]{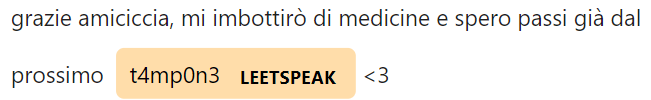}
\caption{\href{https://twitter.com/malakkims/status/1564527881736314881}{\url{Link} to the resource}}
\label{real-example-26}
\end{subfigure}
\hspace{1cm}
\begin{subfigure}[t]{0.46\linewidth}
\includegraphics[width=\linewidth]{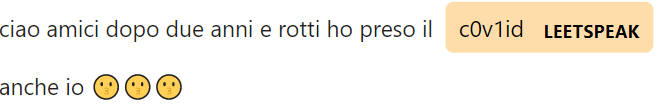}
\caption{\href{https://twitter.com/malakkims/status/1564518968194486274}{\url{Link} to the resource}}
\label{real-example-27}
\end{subfigure}

\end{figure}

\end{document}